%% file: mainv1.tex
\documentclass[10pt,journal,compsoc]{IEEEtran}
\usepackage[ruled,linesnumbered]{algorithm2e}
\usepackage{multirow}
\usepackage[table]{xcolor}
\usepackage{amssymb}
\usepackage{amsfonts}
\usepackage{multicol}
\usepackage{graphicx}
\usepackage{booktabs}
\usepackage[T1]{fontenc}
\usepackage[breaklinks=true,bookmarks=true,pagebackref=false,colorlinks,linkcolor=blue,anchorcolor=red,citecolor=blue,urlcolor=blue]{hyperref}
\usepackage[numbers,sort]{natbib}

\usepackage{comment}
\usepackage{subcaption}
\usepackage{amsmath}
\usepackage{algorithmic}
\usepackage{array}
\usepackage[switch]{lineno}
\usepackage{longtable}
\usepackage{xspace}
\usepackage{url}
\usepackage{overpic}
\usepackage{ragged2e}
\usepackage{framed}
\usepackage{enumitem}
\usepackage{balance}
\input{shortcuts}

\begin{document}

\title{GraphWorld: Long-Horizon Planning with World Models for End-to-End
Autonomous Driving}

% \author{Ziying Song, Caiyan Jia, Lin Liu, Lei Yang, Shengkai Zhang, Feiyang Jia,\\ Fengda Zhao, Peiliang Wu,  Shaoqing Xu, Bin Sun, Yadan Luo

\author{Ziying Song, Caiyan Jia, Lin Liu, Lei Yang, Shengkai Zhang, Feiyang Jia,\\ Fengda Zhao, Peiliang Wu,  Shaoqing Xu, Chen Lv, Yadan Luo
\thanks{Corresponding author: Caiyan Jia.}
\IEEEcompsocitemizethanks{

\IEEEcompsocthanksitem Ziying Song is with the Beijing Key Laboratory of Traffic Data Mining and Embodied Intelligence, School of Computer Science and Technology, Beijing Jiaotong University, Beijing, China, and also with the School of Artificial Intelligence (School of Software), Yanshan University, Qinhuangdao, China.
\IEEEcompsocthanksitem  Caiyan Jia, Lin Liu, Shengkai Zhang, and Feiyang Jia are with Beijing Key Laboratory of Traffic Data Mining and Embodied Intelligence, School of Computer Science and Technology, Beijing Jiaotong University.
\IEEEcompsocthanksitem Lei Yang and Chen Lv are with School of Mechanical and Aerospace Engineering, Nanyang Technological University, Singapore.
\IEEEcompsocthanksitem Fengda Zhao and Peiliang Wu are with the School of Artificial Intelligence (School of Software), Yanshan University, Qinhuangdao, China.

\IEEEcompsocthanksitem Shaoqing Xu is with  University of Macau, China.  
\IEEEcompsocthanksitem Yadan Luo is with  The University of Queensland, Australia. 
}

% \vspace{2cm}

}

\markboth{Submitted to IEEE Transactions on Pattern Analysis and Machine Intelligence
}%
{Song \MakeLowercase{\textit{et al.}}: 
GraphWorld: Long-Horizon Planning with World Models for End-to-End Autonomous Driving
}
% Breaking Imitation Bottlenecks: Reinforced Diffusion Powers Diverse Trajectory Generation
%
%
\IEEEtitleabstractindextext{%
\begin{abstract}
\justifying 
End-to-end autonomous driving has made significant progress by unifying perception, prediction, and planning within a single learning framework, achieving strong performance in short-horizon decision making. However, most existing E2E-AD methods remain confined to short-horizon planning and lack the ability to model long-term temporal dependencies, which severely limits their generalization and security in complex and highly interactive driving scenarios.
In this work, we propose \textbf{GraphWorld}, an E2E-AD framework that explicitly enhances long-horizon planning through latent world modeling. We introduce an Ego-Centric Interaction Graph, which adaptively models critical neighboring agents based on spatial proximity, and propagates relational context to planning queries via cross-node cross-attention. We present a World-State-Conditioned Planning that learns ego-centric latent world representations by modeling interactions between an ego vehicle and surrounding agents. This latent world state captures key interaction dynamics and safety-relevant semantics, and serves as a conditioning signal to guide long-horizon, safety-aware trajectory planning.
Extensive experiments on Bench2Drive, NAVSIMv1/2, and nuScenes demonstrate that GraphWorld significantly reduces collision rates and improves long-horizon planning performance, validating its effectiveness in complex driving environments.
\end{abstract}

}

\maketitle

\IEEEdisplaynontitleabstractindextext

\IEEEpeerreviewmaketitle

\IEEEraisesectionheading{
\section{Introduction}\label{sec:introduction}
}

\IEEEPARstart{E}{n}d-to-end autonomous driving (E2E-AD) has made substantial progress \cite{LiHongyange2etpamisurvey,song2024robustness}, unifying perception, prediction, and planning within a single %unified
framework. Among these tasks, planning serves as the core component of E2E-AD, directly governing the ego vehicle’s decision-making and generating future trajectory outputs for the controlled vehicle \cite{uniad,momad,jiang2023vad,sun2024sparsedrive}.

\begin{figure}[t]
    \centering
    \includegraphics[width=1\linewidth]{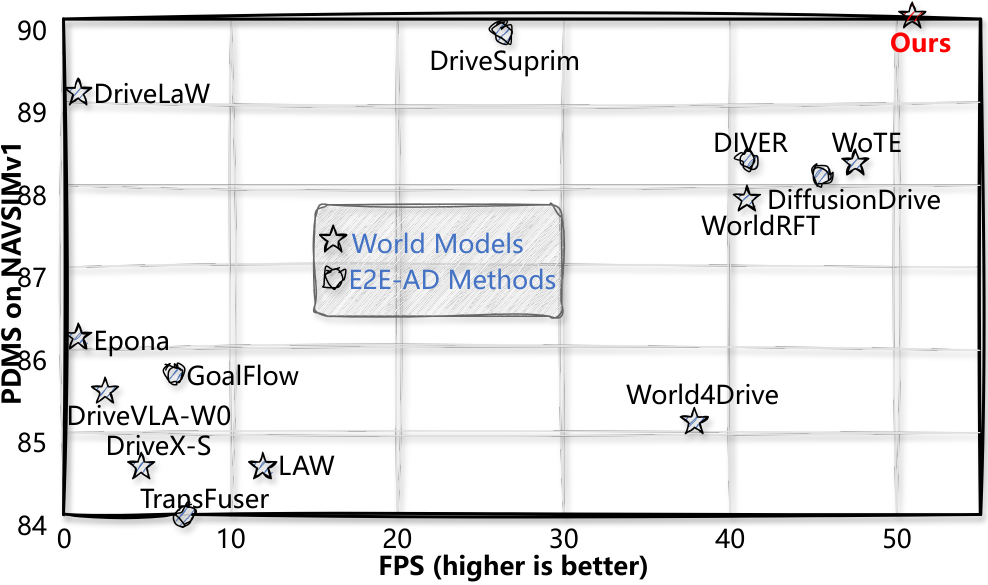}
\caption{Performance comparison of SOTA methods on NAVSIMv1. Compared with representative world-model-based and end-to-end autonomous driving methods, our approach achieves a favorable balance between computational efficiency and prediction accuracy.}
    \label{fig:motivation1}
\end{figure}

\begin{figure*}[t!]
    \centering
    \includegraphics[width=1.0\linewidth]{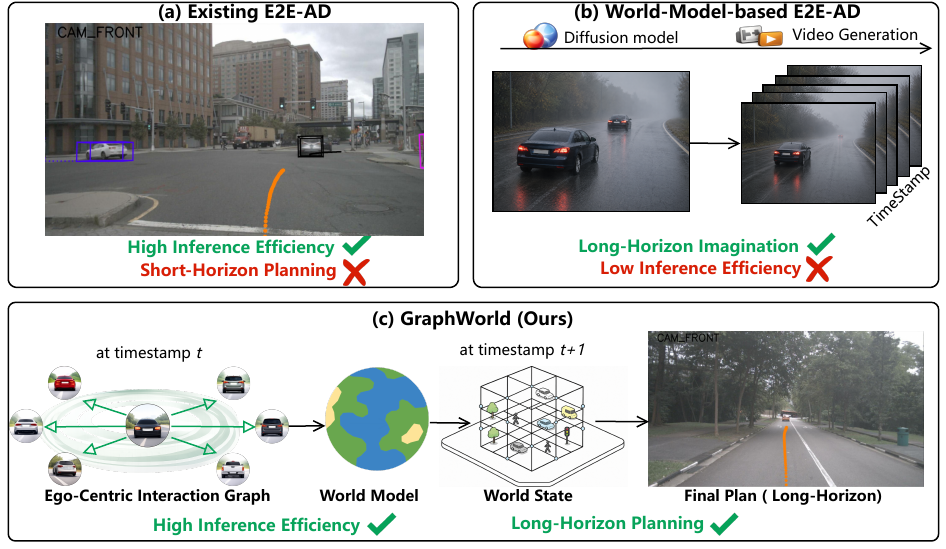}
\caption{\textbf{Motivation of GraphWorld.} 
(a) E2E-AD methods are constrained by short-horizon reasoning, limiting long-term robustness.
(b) World-model approaches enable long-horizon imagination but suffer from low efficiency.
(c) \textbf{GraphWorld} combines \textbf{Ego-Centric Interaction Graph} with \textbf{latent world representations} for \textbf{efficient long-horizon planning}.}
    \label{fig:motivation}
\end{figure*}

E2E-AD has seen continuous innovation in trajectory planning task. Early E2E-AD methods such as UniAD \cite{uniad}, VAD \cite{jiang2023vad}, TransFuser \cite{TransFuser}and UAD \cite{guo2024uad} primarily adopt single-modal trajectory outputs, while subsequent works including VADv2 \cite{chen2024vadv2}, MomAD \cite{momad}, SparseDrive \cite{sun2024sparsedrive}, DiffusionDrive \cite{liao2024diffusiondrive}, and DIVER \cite{diver} extend this line of research toward multi-modal trajectory generation. And anchor-based multi-modal E2E-AD methods, Hydra-MDP \cite{li2024hydra}, GTRS \cite{GTRS}, GoalFlow \cite{xing2025goalflow}, MindDrive \cite{suna2025minddrive}, and GuideFlow \cite{liu2025guideflow} employ more effective trajectory selection. Despite these advances, as shown in Figure \ref{fig:motivation} (a), current E2E-AD methods remain fundamentally constrained by their limited \textit{long-horizon} planning capability. Existing architectures typically process temporal information in a short-sighted manner, lacking mechanisms for modeling and reasoning over extended temporal horizons. 
Long-horizon planning  performance enables multi-step rollout execution at test time, thereby reducing the frequency of online inference and, consequently, lowering the required hardware compute.

Parallel to these E2E-AD developments, world models \cite{feng2025survey_dwm_survey,ding2024understanding_wm_survey,jia2025progressive} have rapidly progressed and %have
demonstrated strong potential in learning environment dynamics and predicting future states through imagination-based inference. Analogous to human cognitive processes, world models possess the ability to simulate and reconstruct possible future scenarios, offering a promising path toward long-horizon reasoning in autonomous driving. When combined with E2E architectures, world models can provide richer temporal context, enhance spatiotemporal understanding, and strengthen anticipation of future events. However, as shown in Figure \ref{fig:motivation} (b), current methods \cite{gao2024vista,hu2023gaia1,wang2024drivedreamer,wang2024driving} that attempt to incorporate world models into E2E-AD largely rely on diffusion-based future scene generation, which is computationally intensive and unsuitable for real-time applications. Moreover, some methods like WoTE \cite{wote} and LAW \cite{li2024enhancing} focus primarily on pixel-level future prediction rather than long-horizon trajectory reasoning, leaving the long-range planning problem essentially underexplored.

In this work, we propose \textbf{GraphWorld}, an E2E-AD framework that explicitly enhances long-horizon planning through latent world modeling. 
Instead of generating future scenes, GraphWorld, as shown in Figure \ref{fig:motivation} (c), learns compact ego-centric relational world representations by modeling interactions between an ego vehicle and its surrounding agents. 
Importantly, rather than relying on explicit multi-step rollout, GraphWorld improves long-horizon planning by learning a latent world state that implicitly encodes future interaction dynamics and safety-relevant evolution. 
To this end, we introduce a dynamically constructed \textit{Ego-Centric Interaction Graph}, which adaptively captures critical neighboring agents based on spatial proximity. 
By focusing on decision-relevant neighbors, ECIG filters out irrelevant interactions and produces a more informative and stable representation. 
Relational context is propagated to planning queries via cross-node cross-attention, enabling effective long-horizon reasoning.
Furthermore, we propose a \textit{World-State-Conditioned Planning} that leverages the inferred ego-centric latent world state to guide trajectory generation. 
The learned world state encodes key interaction dynamics and safety-relevant semantics, and is used to condition multi-modal trajectory queries through a learnable importance network. 
This design allows the planner to emphasize safer and more consistent long-term behaviors while suppressing implausible trajectories.
We first perform E2E pretraining to obtain a stable initialization, and then impose explicit temporal supervision on the world state to encourage temporally consistent and semantically meaningful latent dynamics. 
Extensive experiments on the closed-loop Bench2Drive benchmark and the open-loop nuScenes dataset demonstrate that GraphWorld significantly reduces long-range collision rates and consistently improves long-horizon planning performance, validating its effectiveness in complex environments.
Our contributions are summarized as follows.
\begin{itemize}
    \item We introduce \textbf{GraphWorld}, a novel E2E-AD framework that shifts the paradigm from pixel-space generation to agent-centric relational graph imagination, specifically optimized for long-horizon E2E-AD.
\item We propose a dynamically constructed  Ego-Centric Interaction Graph (ECIG) together with a World-State-Conditioned Planning (WSCP), which jointly extract stable interaction cues and apply them to guide multi-modal trajectory planning via an importance-reweighting network.
\item We evaluate GraphWorld across a comprehensive collection of closed-loop (Bench2Drive, NAVSIM) and open-loop (nuScenes, Adv-nuScenes, nuScenes-C) benchmarks, achieving consistent gains in long-horizon planning and safety. Notably, on the nuScenes 6-second long-horizon planning task, it reduces the average collision rate by 19.5\% over the latent world-model World4Drive \cite{zheng2025world4drive}. Furthermore, it achieves 13.3\% faster speed over DiffusionDrive \cite{liao2024diffusiondrive} on the NAVSIMv1, highlighting its suitability for real-time deployment.
\end{itemize}

\section{Related Work} 
\subsection{End-to-end Autonomous Driving}
E2E-AD methods \cite{uniad,jiang2023vad,chen2024vadv2,sun2024sparsedrive,liao2024diffusiondrive,guo2024uad,xu2024m2da,TransFuser} %maps 
map raw sensor inputs to vehicle control commands or planned trajectories. 
UniAD\cite{uniad} is a comprehensive framework that integrates full-stack driving tasks into a single network. 
% TransFuser\cite{TransFuser} proposes a self-attention–based mechanism to fuse image and LiDAR representations. 
% M2DA \cite{xu2024m2da} introduces a multi-modal fusion transformer with driver-attention modeling. 
VAD \cite{jiang2023vad} presents a fully vectorized scene representation for end-to-end driving. 
% UAD \cite{guo2024uad} adopts an unsupervised pretext task to enhance vision-based autonomous driving. 
MomAD \cite{momad} introduces trajectory and perception momentum to stabilize and refine trajectory outputs. 
% SparseDrive \cite{sun2024sparsedrive} explores sparse scene representation and revisits task design in E2E-AD. 
DiffusionDrive \cite{liao2024diffusiondrive} learns denoising transitions from an anchored Gaussian distribution to multi-modal action distributions. DIVER \cite{diver} combines reinforcement learning with diffusion-based generation to produce diverse and feasible trajectories. Hydra-MDP \cite{li2024hydra} employs a multi-teacher student paradigm. GTRS \cite{GTRS} proposes generalized trajectory scoring for coarse-to-fine trajectory evaluation. GoalFlow \cite{xing2025goalflow} constrains diffusion-generated trajectories through goal-point supervision to address trajectory divergence. 
% MindDrive \cite{suna2025minddrive} integrates high-quality trajectory generation with comprehensive decision reasoning. 
GuideFlow \cite{liu2025guideflow} explicitly models the flow-matching process to mitigate mode collapse and enable flexible conditioning. Despite these advances, existing E2E-AD frameworks remain fundamentally constrained by limited long-horizon planning capability.

\begin{figure*}[t!]
\centering
\includegraphics[width=1.0\linewidth]{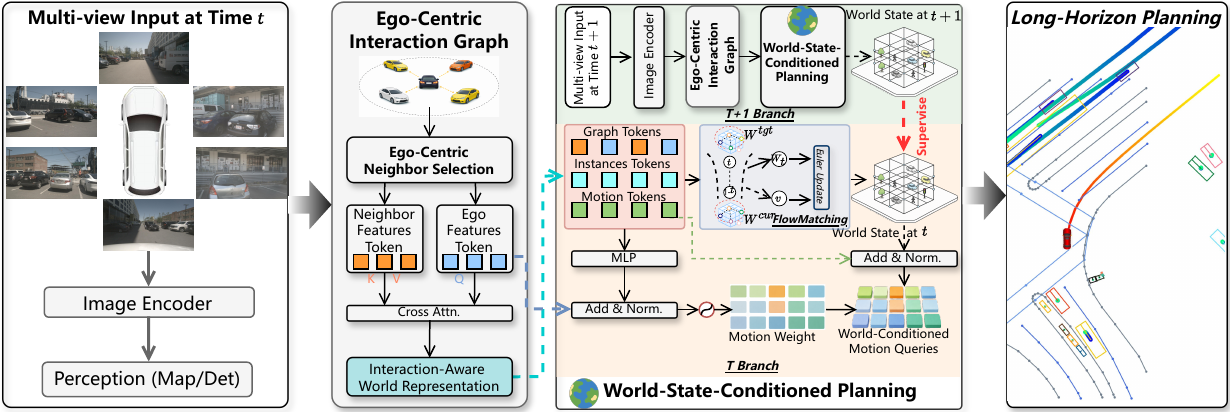}
\caption{\textbf{The proposed long-horizon E2E-AD Framework of GraphWorld.} It first encodes sensor data into instance-level representations, and then constructs an ego-centric latent world model through a dynamically built ECIG that selectively captures critical neighboring agents and their interactions. Based on this interaction-aware world representation, GraphWorld further performs WSCP module by jointly modulating agent motion hypotheses and ego planning queries, enabling interaction-consistent and safety-aware trajectory generation. }
% This design explicitly addresses the challenges of long-horizon reasoning and robustness in complex, highly interactive driving scenarios.  
\label{fig:framework}
\end{figure*}

\subsection{World Models for Autonomous Driving}
The development of world models \cite{liu2024aligning_wm_survey,ha2018recurrent_wm_kaishan,lin2025exploring_wm_survey,liu2025generative_wm_survey,ding2024understanding_wm_survey,lecun2022path} have been rapid. %and 
They are increasingly becoming a key component of autonomous driving, supporting mechanisms for scene dynamic modeling, future event prediction, and downstream decision-making. Driving world models \cite{jia2025progressive,ding2024understanding_wm_survey,feng2025survey_dwm_survey,guan2024world_dwm_survey} have evolved into integrated processes that unify perception, world rule modeling, generation, reconstruction, and multi-task generalization.
A major research direction is the world model optimized through explicit generation. DriveDreamer \cite{wang2024drivedreamer}, Vista \cite{gao2024vista}, UniFuture \cite{liang2025seeing_unifuture}, Epona \cite{zhang2025epona}, and DriveX \cite{shi2025drivex} are capable of high-fidelity video synthesis and controllable policy generation. 
% Furthermore, solutions like OccLLaMA \cite{wei2024occllama}, OccLLM \cite{xu2025occ}, and HERMES \cite{zhou2025hermes} integrate vision, language, and action into the world model paradigm, completing tasks including but not limited to 4D scene prediction, planning, and scene understanding.
Other methods \cite{vasudevan202406_adaptivedriver,qiao2024_ECCV_marlcce} focus on agent-level generation . 
AdaptiveDriver \cite{vasudevan202406_adaptivedriver} predicts the parameters of the agent's motion controller, rather than directly predicting its spatiotemporal trajectory.
Due to their efficiency and abstraction capability, latent world models \cite{zheng2025world4drive,mile-hu2022model,li2025end_wote} have attracted significant attention .
However, most driving world models prioritize explicit generation or reconstruction, limiting their ability to capture spatial semantics and multi-modal driving intentions. Only a few E2E-AD works \cite{li2024enhancing_law,wote} directly leverage latent world models for planning, and their long-term reasoning capabilities are still an area that requires further investigation.

\section{Rethinking Long-Horizon Planning}
\label{sec:rethinking_long_horizon}

In practical autonomous driving systems, planning is typically performed in a receding-horizon manner, where trajectories are iteratively updated at high frequency. 
At each time step $t$, the planner predicts a future trajectory $\tau_t$, but only a short prefix is executed before replanning with updated observations. 
Therefore, long-horizon planning should not be interpreted as executing a fixed long trajectory, but rather as the ability to make foresighted decisions under iterative replanning.

Formally, the planner solves
\begin{equation}
\tau_t^{*} = \arg\max_{\tau_t} p(\tau_t \mid \mathcal{O}_{\leq t}),
\end{equation}
where $\mathcal{O}_{\leq t}$ denotes historical observations. 
Although replanning is frequent, the quality of $\tau_t^{*}$ depends on whether the model can anticipate future interaction risks beyond immediate observations. 
A short-sighted planner may produce locally optimal but globally unsafe decisions under complex multi-agent interactions.

In this work, we do not aim to extend the prediction horizon via explicit multi-step rollout. 
Instead, GraphWorld enhances long-horizon capability by learning a compact ego-centric latent world state
\begin{equation}
\mathbf{W}^{\text{cur}} = f_{\theta}(\mathcal{O}_{\leq t}),
\end{equation}
which encodes interaction dynamics and future-relevant semantics. 
This latent representation enables the planner to reason about the long-term consequences of current actions, even under short-term replanning.

To ensure that the learned representation focuses on decision-relevant structure, we construct an ego-centric interaction graph that selectively models neighboring agents:
\begin{equation}
\mathcal{N} = \left\{ i \mid \lVert \mathbf{p}_i - \mathbf{p}_{\text{ego}} \rVert_2 < \tau \right\}.
\end{equation}
By filtering out irrelevant interactions, the resulting latent world state becomes more stable and informative, which is critical for robust long-horizon reasoning.

Furthermore, instead of performing explicit multi-step prediction, we model world evolution in a latent space via a continuous transformation between current and target states:
\begin{equation}
\mathbf{W}(t) = (1-t)\mathbf{W}^{\text{cur}} + t\mathbf{W}^{\text{tgt}}, \quad t \sim \mathcal{U}(0,1),
\end{equation}
which avoids error accumulation commonly observed in autoregressive rollout while preserving future-aware structure.

Regarding evaluation, although we follow standard benchmarks with prediction horizons up to 6 seconds, performance at later timestamps (e.g., 5--6s) reflects robustness under compounding uncertainty. 
Improvements in this regime indicate stronger capability in handling long-term interactions and safety-critical scenarios.

Overall, we argue that long-horizon planning should be understood as the ability to learn future-aware representations that support robust decision-making under iterative replanning, rather than merely extending the prediction horizon.

\section{Method}
GraphWorld is an E2E-AD framework designed to enhance long-horizon planning through latent world modeling. As shown in Figure~\ref{fig:framework}, GraphWorld constructs an ego-centric latent world representation from multi-agent instance features extracted from perception by dynamically modeling interactions between the ego vehicle and surrounding agents. The framework consists of two key components:
(1) an \textbf{Ego-Centric Interaction Graph (ECIG)} that encodes ego--agent interactions in a structured relational space, and
(2) a \textbf{World-State-Conditioned Planning (WSCP)} that leverages the inferred world state to enable long-horizon, safety-aware trajectory planning.

\subsection{Ego-Centric Interaction Graph (ECIG)}
\label{sec:ecig}
We introduce an ECIG that explicitly encodes salient ego--agent interactions and injects structured relational context into ego planning queries via cross-attention, enabling effective long-horizon reasoning.

\noindent \textbf{Ego-centric Neighbor Selection.}
At each planning step, an ego vehicle and its surrounding agents are represented as graph nodes. Let
$
\mathbf{F} = \{\mathbf{f}_1, \dots, \mathbf{f}_N, \mathbf{f}_{\text{ego}}\}, 
\mathbf{f}_i \in \mathbb{R}^{C}
$
denote instance-level features at a single planning step for $N$ selected agents and the ego vehicle. The ego node is explicitly included as the center of interaction modeling. To construct a sparse interaction graph, ECIG dynamically selects a fixed number of neighboring agents based on spatial proximity. Let $\mathbf{p}_i \in \mathbb{R}^2$ denote the predicted 2D center of agent $i$ and $\mathbf{p}_{\text{ego}}$ the ego center. We define an ego-centric neighbor set
$
\mathcal{N} = \left\{ i \;\middle|\; \lVert \mathbf{p}_i-\mathbf{p}_{\text{ego}} \rVert_2 < \tau \right\},
$
where $\tau$ is a distance threshold. 
% The neighbor indices are then padded or trimmed to a fixed size $K$:
% \begin{equation}
% \mathcal{N}_K = \textsc{PadOrTrim}(\mathcal{N}, K).
% \end{equation}
% If $\mathcal{N}$ is empty, a fallback index is inserted before padding to ensure the construction of a valid graph.
As the cardinality of the neighbor set may vary across scenes, we apply a \textsc{PadOrTrim} operation to obtain a fixed-size $K$-neighbor set:
\begin{equation}
\mathcal{N}_K = \textsc{PadOrTrim}(\mathcal{N}, K).
\end{equation}
\noindent \textbf{Ego-centric Interaction Graph.}
The resulting interaction graph is defined as $\mathcal{G} = (\mathcal{V}, \mathcal{E})$ with
$\mathcal{V} = \{\text{ego}\} \cup \mathcal{N}_K$, where directed edges $\mathcal{E}$ connect each neighbor node to the ego node, forming an ego-centered star topology.
Let $ \mathbf{Q}_{\text{ego}} \in \mathbb{R}^{M \times C}$
denote $M$ multi-modal ego planning queries, and
$ \mathbf{F}_{\text{nbr}} = \{\mathbf{f}_i \mid i \in \mathcal{N}_K\} \in \mathbb{R}^{K \times C} $
denote neighbor features. ECIG propagates interaction information through cross-attention:
\begin{equation}
\tilde{\mathbf{Q}}_{\text{ego}} =
\textsc{CrossAttn}(\mathbf{Q}_{\text{ego}}, \mathbf{F}_{\text{nbr}}, \mathbf{F}_{\text{nbr}}),
\end{equation}
where ego queries attend to neighboring agent features, enabling each planning hypothesis to selectively aggregate interaction-aware context.

\noindent \textbf{Interaction-aware World Representation.}
To construct a structured and temporally grounded latent world state, we condition world representation learning on multi-source contextual cues, including interaction structure, motion history, and static map semantics. Specifically, given interaction-enhanced ego queries $\tilde{\mathbf{Q}}_{\text{ego}}$, neighbor features $\mathbf{F}_{\text{nbr}}$, historical ego-agent states $\mathbf{H}_{t}=\{\mathbf{h}_{t-L},\dots,\mathbf{h}_{t}\}$ of $L$-step horizon, and local map embeddings $\mathbf{M}$, we first model temporal dynamics using a recurrent world encoder:
\begin{equation}
\mathbf{s}_t = \textsc{GRU}\!\left(\mathbf{s}_{t-1},\; \textsc{Concat}(\mathbf{H}_{t}, \mathbf{M}) \right),
\end{equation}
where $\mathbf{H}_t$ is constructed by temporally stacking the motion states (e.g., position, velocity, and heading) of the ego vehicle and its neighbors over the past (L) frames,$\mathbf{s}_t$ denotes the temporally accumulated world context at time $t$.
Here, map features $\mathbf{M}$ are first pooled into a compact embedding before concatenation with the historical state sequence.
We then integrate interaction-aware information via cross-attention:
\begin{equation}
\mathbf{c}_{\text{int}} =
\textsc{CrossAttn}\!\left(
\mathbf{s}_t,\;
\tilde{\mathbf{Q}}_{\text{ego}} \cup \mathbf{F}_{\text{nbr}},\;
\tilde{\mathbf{Q}}_{\text{ego}} \cup \mathbf{F}_{\text{nbr}}
\right).
\end{equation}
Based on the resulting context, we update node-level latent world states through conditioned residual injection:
\begin{equation}
\mathbf{w}_i = \mathbf{f}_i + \phi_a^{\text{int}}(\mathbf{c}_{\text{int}}, \mathbf{r}_i), \quad
\mathbf{w}_{\text{ego}} = \mathbf{f}_{\text{ego}} + \phi_e^{\text{int}}(\mathbf{c}_{\text{int}}),
\end{equation}
where the ego-centric relational geometry of agent $i$ is denoted by $\mathbf{r}_i=\textsc{MLP}\left([\mathbf{f}_i,\mathbf{p}_i-\mathbf{p}_{\text{ego}}]\right)$, and lightweight conditional projection networks are $\phi_a^{\text{int}}(\cdot)$, $\phi_e^{\text{int}}(\cdot)$. The resulting latent world state $\mathbf{W}=\{\mathbf{w}_1,\dots,\mathbf{w}_K,\dots,\mathbf{w}_N,\mathbf{w}_{\text{ego}}\}$ captures temporally consistent ego-agent interaction dynamics under map constraints and conditions downstream long-horizon planning. Here, $K$ denotes the number of selected neighbors and $N$ denotes the total number of agent anchors (set to $900$ in nuScenes); the anchors $\mathbf{w}_{K+1:N}=\mathbf{0}$ are zero-padded.
% The resulting latent world state $\mathbf{W}=\{\mathbf{w}_1,\dots,\mathbf{w}_K,\dots,\mathbf{0},\dots,\mathbf{w}_N,\mathbf{w}_{\text{ego}}\}$ captures temporally consistent ego--agent interaction dynamics under map constraints and serves as a structured conditioning signal for downstream long-horizon, safety-aware planning. Here, $K$ denotes the number of selected neighbor agents, and $N$ denotes the total number of agent anchors (set to $900$ in nuScenes); unused anchors are zero-padded, i.e., $\mathbf{w}_{K+1:N}=\mathbf{0}$.

\subsection{World-State-Conditioned Planning (WSCP)}
\label{sec:wscp}

Given the interaction-aware world representation inferred by ECIG, we propose a \emph{WSCP} module to enable long-horizon, safety-aware trajectory generation.
Instead of decoding trajectories solely from planning queries, WSCP explicitly conditions both agent motion hypotheses and ego planning on the latent world dynamics, allowing interaction-consistent reasoning over extended horizons.

\noindent \textbf{Flow-Matching for Dynamic World-State.}
To explicitly model long-horizon world evolution in a principled continuous-time manner, we formulate world-state dynamics using a conditional \emph{Flow-Matching} framework \cite{lipman2022flow}. Rather than performing discrete autoregressive updates or iterative diffusion sampling, Flow-Matching directly learns a time-dependent velocity field that transports latent world states along interaction-consistent trajectories. 

For clarity, we distinguish between a \emph{current} latent world state
$\mathbf{W}^{\text{cur}}$ inferred from present observations, and a
\emph{conditioned target} world state $\mathbf{W}^{\text{tgt}}$ constructed from
future agent motion hypotheses, ego planning context, and static map semantics.
The current world state is given by
$
\mathbf{W}^{\text{cur}} = \{\mathbf{w}_1,\dots,\mathbf{w}_N,\mathbf{w}_{\text{ego}}\},
$
as inferred by the ECIG.

The target world state $\mathbf{W}^{\text{tgt}}$ is constructed by conditioning on
multi-modal agent motion hypotheses, ego planning context, and optional map embeddings,
and is used solely as a supervision signal for world-state evolution rather than an
explicit rollout target. Formally, we define
\begin{equation}
\mathbf{W}^{\text{tgt}}
=
\phi_{\text{tgt}}\!\left(
\textsc{Concat}\!\left(
\bar{\mathbf{Q}}_{\text{motion}},\;
\bar{\mathbf{Q}}_{\text{ego}},\;
\mathbf{M}
\right)
\right),
\end{equation}
where the mode-aggregated agent motion latents and ego planning representations are denoted by $\bar{\mathbf{Q}}_{\text{motion}}=\textsc{Mean}_{m}(\mathbf{Q}_{\text{motion}})$
and $\bar{\mathbf{Q}}_{\text{ego}}=\textsc{Mean}_{m}(\tilde{\mathbf{Q}}_{\text{ego}})$, respectively.
$\mathbf{M}$ denotes optional map embeddings, and $\phi_{\text{tgt}}(\cdot)$ is a
lightweight projection network mapping contextual cues to the latent world space.

We define a continuous interpolation between the current and target world states:
\begin{equation}
\mathbf{W}(t) = (1 - t)\,\mathbf{W}^{\text{cur}} + t\,\mathbf{W}^{\text{tgt}},
\quad t \sim \mathcal{U}(0,1),
\end{equation}
where $t$ denotes a randomly sampled continuous time.
The corresponding ground-truth velocity along this path is
\begin{equation}
\mathbf{v}^\star(t) = \frac{d\mathbf{W}(t)}{dt}
= \mathbf{W}^{\text{tgt}} - \mathbf{W}^{\text{cur}} .
\end{equation}
We parameterize a conditional velocity field
\begin{equation}
\mathbf{v}_\theta(t) =
f_\theta\!\left(
\mathbf{W}(t),\;
t,\;
\bar{\mathbf{Q}}_{\text{ego}},\;
\bar{\mathbf{F}}_{\text{nbr}},\;
\mathbf{M}
\right),
\end{equation}
where $f_\theta(\cdot)$ is a lightweight neural network,
$\bar{\mathbf{Q}}_{\text{ego}}=\textsc{Mean}_m(\tilde{\mathbf{Q}}_{\text{ego}})$
aggregates multi-modal ego planning hypotheses,
$\bar{\mathbf{F}}_{\text{nbr}}=\textsc{Mean}(\mathbf{F}_{\text{nbr}})$
summarizes the surrounding agent context,
and $\mathbf{M}$ denotes optional local map embeddings.
In this way, the learned flow is explicitly conditioned on ego--agent interactions
and static scene constraints.

Training is performed using the Flow-Matching objective:
\begin{equation}
\mathcal{L}_{\text{FM}} =
\mathbb{E}_{t,\mathbf{W}^{\text{cur}},\mathbf{W}^{\text{tgt}}}
\left[
\left\|
\mathbf{v}_\theta(t) - \mathbf{v}^\star(t)
\right\|_2^2
\right],
\end{equation}
which enforces the predicted velocity field to match the true transport direction
at arbitrary intermediate world states.  At inference time, the learned velocity field defines a continuous-time dynamical system:
\begin{equation}
\frac{d\mathbf{W}(t)}{dt} = \mathbf{v}_\theta(t),
\end{equation}
which is approximated using a single-step Euler update:
\begin{equation}
\mathbf{W}_t
\;\leftarrow\;
\mathbf{W}_t
+
\Delta t \cdot \mathbf{v}_\theta(t).
\end{equation}
This Flow-Matching formulation enables stable and interaction-consistent refinement of the latent world state, serving as a temporally grounded conditioning signal for long-horizon, safety-aware planning. And during generation, the number of sampling steps is set to 2.
% and在生成过程中，我们的采样步长被设置为25

% \noindent \textbf{World-conditioned motion query modulation.}
% Let $\mathbf{W}_{t}=\{\mathbf{W}_{1:N},\mathbf{W}_{\text{ego}}\}$ denote the refined latent world state at time $t$.
% To inject long-horizon world dynamics into trajectory hypotheses, we condition
% agent motion queries on the updated agent world state:
% \begin{equation}
% \mathbf{C} = \phi_c(\mathbf{W}_{1:N}) \in \mathbb{R}^{B \times N \times C},
% \end{equation}
% \begin{equation}
% \mathbf{Q}'_{\text{motion}}
% =
% \textsc{LN}\!\left(
% \mathbf{Q}_{\text{motion}} + \mathbf{C}[:, :, \text{None}, :]
% \right).
% \end{equation}

\noindent \textbf{Importance Reweighting for Motion Modeling.}

Let $\mathbf{W}_{t}=\{\mathbf{W}_{1:N},\mathbf{W}_{\text{ego}}\}$ denote the refined latent world state at time $t$.
To inject long-horizon world dynamics into trajectory hypotheses, we condition
agent motion queries on the updated agent world state:
\begin{equation}
\mathbf{C} = \phi_c(\mathbf{W}_{1:N}) \in \mathbb{R}^{B \times N \times C},
\end{equation}
\begin{equation}
\mathbf{Q}'_{\text{motion}}
=
\textsc{LN}\!\left(
\mathbf{Q}_{\text{motion}} + \mathbf{C}[:, :, \text{None}, :]
\right).
\end{equation}
Long-horizon planning is sensitive to the reliability of surrounding-agent predictions.
We therefore perform mode-wise importance re-weighting using ego-conditioned world dynamics.
Specifically,
\begin{equation}
\boldsymbol{\alpha}
=
\sigma\!\left(
f_{\text{imp}}\!\left(
\textsc{Concat}\!\left(
\mathbf{Q}'_{\text{motion}} + \mathbf{W}_{1:N}^{\uparrow},
\mathbf{f}_{\text{ego}}^{\uparrow}
\right)
\right)
\right),
\end{equation}
where $f_{\text{imp}}(\cdot)$ denotes a lightweight two-layer MLP, $\sigma(\cdot)$ is the sigmoid function and $\uparrow$ denotes broadcast
along agent and mode dimensions.
The re-weighted motion queries are
\begin{equation}
\hat{\mathbf{Q}}_{\text{motion}}
=
\boldsymbol{\alpha} \odot \mathbf{Q}'_{\text{motion}},
\end{equation}
where $\boldsymbol{\alpha}$ reflects the world-conditioned reliability of each motion mode.
\noindent \textbf{World-to-Ego Feedback for Planning Queries.}
The refined world state further conditions ego planning via residual modulation:
\begin{equation}
\mathbf{Q}'_{\text{plan}} = \mathbf{Q}_{\text{plan}} + \mathbf{W}_{\text{ego}}.
\end{equation}

\noindent \textbf{Multi-modal Motion and Ego Trajectory Decoding.}
Finally, the re-weighted agent motion queries and world-conditioned ego planning
queries are jointly decoded by task-specific heads:
\begin{equation}
\left\{
\begin{aligned}
\{\boldsymbol{\pi}_{n,m}, \hat{\mathbf{Y}}_{n,m}\}
&= f_{\text{motion}}\!\left(\hat{\mathbf{Q}}_{\text{motion}}\right), \\
\{\boldsymbol{\pi}^{\text{ego}}_{m}, \hat{\mathbf{Y}}^{\text{ego}}_{m}\}
&= f_{\text{plan}}\!\left(\mathbf{Q}'_{\text{plan}}\right),
\end{aligned}
\right.
\end{equation}
where $\hat{\mathbf{Y}}_{n,m} \in \mathbb{R}^{T \times 2}$ denotes the $m$-th predicted
future trajectory of agent $n$, and $\boldsymbol{\pi}_{n,m}$ is its corresponding
mode probability.
Similarly, $\hat{\mathbf{Y}}^{\text{ego}}_{m} \in \mathbb{R}^{T \times 2}$ denotes the
$m$-th ego trajectory hypothesis, with $\boldsymbol{\pi}^{\text{ego}}_{m}$ indicating its corresponding confidence score.
% All probabilities are normalized across modes and reflect the relative plausibility of each trajectory under the inferred world dynamics.

\begin{table*}[t]
\scriptsize
\centering
\caption[]{Planning results for \textbf{6-second long-horizon planning} on the \textbf{nuScenes} validation set.}
\renewcommand\arraystretch{0.7}
\setlength{\tabcolsep}{1.99mm}
  \resizebox{\linewidth}{!}{
\begin{tabular}{l ccccccc  ccccccc }
\toprule
\multirow{2}{*}{$\operatorname{Method}$} & \multicolumn{7}{c}{$\operatorname{L2\ (m)}\downarrow$} & \multicolumn{7}{c}{$\operatorname{Col.\ Rate\ (\%)}\downarrow$}  \\
\cmidrule(lr){2-8} \cmidrule(lr){9-15}
& 1s & 2s & 3s & 4s & 5s & 6s& $\operatorname{Avg.}$ & 1s & 2s & 3s & 4s & 5s & 6s &$\operatorname{Avg.}$ \\
\midrule
$\operatorname{UniAD}$~\cite{uniad} &0.47& 0.91& 1.35 &1.91& 2.47 &3.07&\cellcolor{red!5}1.70& 0.25& 0.36& 0.61 & 0.99 &1.64& 2.51&\cellcolor{red!5}1.06 \\
$\operatorname{SparseDrive}$~\cite{sun2024sparsedrive} &0.43& 0.87& 1.23 &1.75& 2.32 &2.95&\cellcolor{red!5}1.59& 0.19& 0.31& 0.56 & 0.87 &1.54& 2.33&\cellcolor{red!5}0.97 \\
$\operatorname{MomAD}$~\cite{momad}&0.41& 0.85& 1.13 &1.67& 1.98& 2.45&\cellcolor{red!5}1.42&0.17& 0.30& 0.54 & 0.83& 1.43 &2.13&\cellcolor{red!5}0.90\\
$\operatorname{LAW}$~\cite{li2024enhancing_law}&0.40& 0.87& 1.16 &1.71& 2.03& 2.61&\cellcolor{red!5}1.46&0.19& 0.33& 0.57 & 0.86& 1.51 &2.31&\cellcolor{red!5}0.96\\
$\operatorname{Epona}$~\cite{zhang2025epona}&0.39& 0.91& 1.17 &1.73& 2.02& 2.75&\cellcolor{red!5}1.50&0.14& 0.18& 0.45 & 0.74& 1.48 &2.23&\cellcolor{red!5}0.87\\
$\operatorname{World4Drive}$~\cite{zheng2025world4drive}&0.42& 0.92& 1.21 &1.75& 2.06& 2.79&\cellcolor{red!5}1.53&0.16& 0.20& 0.47 & 0.76& 1.50 &2.14&\cellcolor{red!5}0.87\\
\cellcolor{red!5}$\operatorname{GraphWorld (Ours)}$&\cellcolor{red!5}\textbf{0.38}& \cellcolor{red!5}\textbf{0.70} & \cellcolor{red!5}\textbf{1.12} &\cellcolor{red!5}\textbf{1.64} &\cellcolor{red!5}\textbf{1.88} & \cellcolor{red!5}\textbf{2.29}&\cellcolor{red!5}\textbf{1.34}&\cellcolor{red!5}\textbf{0.03}& \cellcolor{red!5}\textbf{0.06}&\cellcolor{red!5}\textbf{0.25}& \cellcolor{red!5}\textbf{0.65}& \cellcolor{red!5}\textbf{1.29} &\cellcolor{red!5}\textbf{1.95}&\cellcolor{red!5}\textbf{0.70}\\
\bottomrule
\end{tabular}}
\label{tab_nuscenes_planning_6s}
\end{table*}

\subsection{Two-Stage Temporal World-State Supervision}
\label{sec:training}
To encourage the world state to learn a genuine representation of future state evolution, we adopt a two-stage training strategy. In the first stage, we perform standard end-to-end training of the overall model. In the second stage, we apply explicit supervision to the world state of the world model.
\noindent \textbf{Stage I: Single-step world state learning.}
In the first stage, the model is trained using inputs at time $t$ only.
Given multi-view observations at time $t$, GraphWorld infers a latent world state
$\mathbf{W}_t$ through the ECIG and WSCP modules.
The network is optimized with standard perception, motion prediction, and ego planning losses, allowing the world state to encode interaction-aware and safety-relevant scene semantics.

\noindent \textbf{Stage II: Temporal world state consistency learning.}
In the second stage, we introduce temporal supervision by leveraging observations at time $t{+}1$.
Specifically, the model predicts the next-step world state $\mathbf{W}_{t+1}$ from inputs at time $t{+}1$,
while the world state inferred at time $t$ is encouraged to be temporally consistent with $\mathbf{W}_{t+1}$.
We enforce this consistency using an $\ell_2$ regression loss:
\begin{equation}
\mathcal{L}_{\text{world}} =
\left\| \mathbf{W}_t - \textsc{StopGrad}\!\left(\mathbf{W}_{t+1}\right) \right\|_2^2,
\end{equation}
where $\textsc{StopGrad}(\cdot)$ prevents gradient backpropagation through the $t{+}1$ branch.

\noindent \textbf{Training objective.}
The final training loss is a weighted combination of task-specific losses and the temporal world-state consistency loss:
\begin{equation}
\mathcal{L} =
\mathcal{L}_{\text{percep}} +
\mathcal{L}_{\text{motion}} +
\mathcal{L}_{\text{plan}} +
\lambda\, \mathcal{L}_{\text{world}},
\end{equation}
where $\lambda$ balances temporal consistency and task performance.

\begin{table*}[t]
\scriptsize
\centering
\addtolength{\tabcolsep}{0.1pt}
\caption{$\operatorname{ \textbf{Open-Loop}}$, $\operatorname{\textbf{Closed-Loop}}$ results and $\operatorname{Multi-Ability}$ results on $\operatorname{\textbf{Bench2Drive}}$ (V0.0.3) under base training set.
`mmt' refers multi-mode trajectory variant of $\operatorname{VAD}$ and $^\dagger$ denotes the re-implementation. * denotes expert feature distillation. `DS' denotes Driving Score. `SR' denotes Success Rate. `Effi' denotes Efficiency. `Comf' denotes Comfortness. `Merg.' denotes Merging.  `Overta.' denotes Overtaking. `Emerge.' denotes Emergency Brake. 
}
  \renewcommand\arraystretch{0.7}
  \tabcolsep=0.1mm %%%%%%%%%
  \resizebox{\linewidth}{!}{
  \begin{tabular}{l l cc cccc cccccc}
    \toprule
   \multirow{2}{*}{$\operatorname{Method}$}& \multirow{2}{*}{$\operatorname{Venue}$}&  \multicolumn{1}{c}{$\operatorname{Open-loop\ Metric}$}    & \multicolumn{4}{c}{$\operatorname{Closed-loop\ Metric}$}& \multicolumn{6}{c}{$\operatorname{Multi-Ability (\%) \uparrow}$}  \\
   \cmidrule(lr){3-3}\cmidrule(lr){4-7}\cmidrule(lr){8-13}
   &&$\operatorname{Avg. L2}\downarrow$&$\operatorname{DS}\uparrow$
   & $\operatorname{SR\ (\%)}\uparrow$ & $\operatorname{Effi}\uparrow$  & $\operatorname{Comf}\uparrow$ & $\operatorname{Merg.}\uparrow$
   & $\operatorname{Overta.}$& $\operatorname{Emerge.}$& $\operatorname{Give\:Way}$ &$\operatorname{Traffic\:Sign}$& $\operatorname{Mean}$ \\
   
\midrule
% TCP-traj* \cite{e_2_e_TCP}&NeurIPS2022&1.70& 59.90& 30.00& 76.54& 18.08& 8.89& 24.29& \textbf{51.67}& 40.00& 46.28& 34.22\\
ThinkTwice*  \cite{thiktwice}&CVPR2023&0.95&  62.44& 31.23 &69.33& 16.22& 27.38&18.42& 35.82& \textbf{50.00}& 54.23& 37.17\\
DriveAdapter* \cite{jia2023driveadapter}&ICCV2023 & 1.01& 64.22& 33.08 &70.22& 16.01& 28.82&26.38& 48.76& \textbf{50.00}& 56.43& 42.08\\
DriveDPO \cite{shang2025drivedpo}&   NeurIPS2025&-&  62.02& 30.62& -& -& - &- &- &-&- &-\\
Raw2Drive \cite{Raw2Drive}& NeurIPS2025 &-& 71.36& \textbf{50.24}&-&-& \textbf{43.35} & \textbf{51.11}& \textbf{60.00} &\textbf{50.00} &62.26 &\textbf{53.34}\\
DriveTrans* \cite{jia2025drivetransformer}&ICLR2025&\textbf{0.62} & 63.46& 35.01 &\textbf{100.64}& 20.78& 17.57& 35.00& 48.36& 40.00& 52.10& 38.60\\
WoTE* \cite{wote}& ICCV2025&-  &61.71& 31.36 &-& -& -& -& -& -& -& -\\
\midrule
${\operatorname{ThinkTwice}_{\operatorname{mmt}}}^{*\dagger }$ \cite{thiktwice}&CVPR2023&0.93&  63.34& 33.23 &71.56& 18.32& 31.31&21.23& 38.33& \textbf{50.00}& 57.45& 39.66\\
\cellcolor{red!5}${\operatorname{GraphWorld}}$ (Ours)&\cellcolor{red!5}& \cellcolor{red!5}0.81& \cellcolor{red!5}\textbf{76.71}
& \cellcolor{red!5}44.22
& \cellcolor{red!5}80.48
& \cellcolor{red!5}\textbf{29.74}
& \cellcolor{red!5}42.85
& \cellcolor{red!5}32.57
& \cellcolor{red!5}49.09
& \cellcolor{red!5}\textbf{50.00}
& \cellcolor{red!5}\textbf{67.16}
& \cellcolor{red!5}48.33\\

\midrule
\midrule
% $\operatorname{AD-MLP}$ \cite{ADmlp}&Arxiv 2023&3.64& -&  18.05& 0.00& 48.45& 22.63& 0.00& 0.00& 0.00& 0.00& 4.35& 0.87\\
UniAD-Base \cite{uniad}&CVPR2023&0.73 & 45.81& 16.36& 129.21& 43.58& 14.10& 17.78& 21.67& 10.00& 14.21& 15.55\\
$\operatorname{VAD}$ \cite{jiang2023vad}&ICCV2023& 0.91& 42.35& 15.00& 157.94& 46.01&8.11& 24.44& 18.64& \textbf{20.00}& 19.15& 18.07 \\
GenAD \cite{zheng2024genad}&ECCV2024 &-&44.81& 15.90& -& -& -& -& -& -& -& -\\
% $\operatorname{MomAD(VAD)}$ \cite{momad}&CVPR2025&0.87 & 45.35& 17.44& 162.09& 49.34&9.99& 26.31& 20.07& \textbf{20.00}& 20.23&19.32 \\
$\operatorname{MomAD(SD)}$ \cite{momad}&CVPR2025&0.82  &47.91& 18.11& 174.91 &51.20&13.21& 21.02& 18.01& \textbf{20.00}& 21.07&18.66  \\

\midrule
${\operatorname{VAD}_{\operatorname{mmt}}}^{\dagger }$ \cite{jiang2023vad} &ICCV 2023&0.89&  42.87& 15.91& 158.12& 47.22&9.43& 25.31& 19.91& \textbf{20.00}& 20.09&18.95   \\
\cellcolor{red!5}${\operatorname{GraphWorld}}$ (Ours)&\cellcolor{red!5}&\cellcolor{red!5}0.81 & \cellcolor{red!5}50.78& \cellcolor{red!5}22.59& \cellcolor{red!5}168.11& \cellcolor{red!5}54.04& \cellcolor{red!5}16.84& \cellcolor{red!5}\textbf{32.31}& \cellcolor{red!5}\textbf{28.46}& \cellcolor{red!5}\textbf{20.00}& \cellcolor{red!5}\textbf{27.61}& \cellcolor{red!5}25.04 \\
\midrule
${\operatorname{SparseDrive}}^{\dagger}$ \cite{sun2024sparsedrive}&ICRA2025&0.87& 44.54& 16.71& 170.21& 48.63&12.18& 23.19& 17.91& \textbf{20.00}& 20.98& 18.85 \\
\cellcolor{red!5}${\operatorname{GraphWorld}}$ (Ours) &\cellcolor{red!5}&\cellcolor{red!5}0.79&\cellcolor{red!5}\textbf{51.55}& \cellcolor{red!5}\textbf{25.47}& \cellcolor{red!5}\textbf{181.12}& \cellcolor{red!5}\textbf{56.59}& \cellcolor{red!5}\textbf{18.74}& \cellcolor{red!5}31.66& \cellcolor{red!5}25.30& \cellcolor{red!5}\textbf{20.00}& \cellcolor{red!5}26.66& \cellcolor{red!5}\textbf{24.47}\\

\bottomrule
\end{tabular}
}
\label{tab_b2d}
\end{table*}

\begin{table}[t]
\centering
  \caption{\textcolor{black}{Comparison on planning-oriented \textbf{NAVSIMv1} navtest split with $\operatorname{\textbf{Closed-Loop}}$ metrics. `mmt' refers multi-modal trajectory variant of $\operatorname{TransFuser}$ and $^*$ denotes the re-implementation. }}
\renewcommand\arraystretch{0.7}
  \tabcolsep=0.6mm %%%%%%%%%
  \resizebox{\linewidth}{!}{
    \begin{tabular}{l c c c c c c}
    \toprule
    \multicolumn{1}{l}{Method}&  NC$\uparrow$& DAC$\uparrow$& TTC$\uparrow$& Conf.$\uparrow$& EP$\uparrow$& PDMS$\uparrow$ \\
            \midrule
            \rowcolor{cyan!5}\multicolumn{7}{c}{\textit{E2E-based Methods}} \\
            % Ego Status MLP &  93.0 & 77.3 & 83.6 & 100 & 62.8 & 65.6 \\
            VADv2~\cite{chen2024vadv2} &  97.2 & 89.1 & 91.6 & 100 & 76.0 & 80.9 \\
            TransFuser~\cite{TransFuser} &  97.7 & 92.8 & 92.8 & 100 & 79.2 & 84.0 \\
            ${\operatorname{TransFuser}_{\operatorname{mmt}}}^{*}$ \cite{TransFuser}&96.2& 95.4& 90.7& 100& 80.7& 85.1 \\ 
            UniAD~\cite{uniad} &  97.8 & 91.9 & 92.9 & 100 & 78.8 & 83.4 \\
            PARA-Drive~\cite{paradrive} &  97.9 & 92.4 & 93.0 & 99.8 & 79.3 & 84.0 \\
            DRAMA~\cite{yuan2024drama} &  98.0 & 93.1 & 94.8 & 100 & 80.1 & 85.5 \\
            GoalFlow~\cite{xing2025goalflow}  & 98.3 & 93.8 & 94.3 & 100 & 79.8 & 85.7 \\
            Hydra-MDP~\cite{li2024hydra} &  98.3 & 96.0 & 94.6 & 100 & 78.7 & 86.5 \\
            % ARTEMIS~\cite{feng2025artemis} &  98.3 & 95.1 & 94.3 & 100 & 81.4 & 87.0 \\
            FUMP~\cite{liu2025fump} &  98.1 & 96.2 & 94.2 & 100 & 82.0 & 87.8 \\
            DiffusionDrive~\cite{liao2024diffusiondrive} & 98.2 & 96.2 & 94.7 & 100 & 82.2 & 88.1 \\
            DIVER~\cite{diver} &  98.5 & 96.5 & 94.9 & 100 & 82.6 & 88.3 \\
            DriveSuprim~\cite{yao2025drivesuprim} &  97.8 & 97.3 & 93.6 & 100 & 86.7 & 89.9 \\
            GoalFlow~\cite{xing2025goalflow}  &  98.4 & 98.3 & 94.6 & 100 & 85.0 & 90.3 \\
            ReCogDrive-IL\cite{li2025recogdrive}&98.1& 94.7 &94.2& 100& 80.9 &86.5\\
            \midrule
            \rowcolor{cyan!5}\multicolumn{7}{c}{\textit{VLA-based Methods}} \\
           AutoVLA ~\cite{zhou2025autovla} & 98.4 & 95.6 & 98.0 & 99.9 & 81.9 & 89.1 \\
            Recogdrive~\cite{li2025recogdrive} &  98.2 & 97.8 & 95.2 & 99.8 & 83.5 & 89.6 \\
            DriveVLA-W0~\cite{li2025drivevla} &  98.7 & 99.1 & 95.3 & 99.3 & 83.3 & 90.2 \\
            DriveWorld-VLA~\cite{liu2026driveworld} &  99.1 & 98.2 & 96.1 & 100 & 85.9 & 91.3 \\
            \midrule
            \rowcolor{cyan!5}\multicolumn{7}{c}{\textit{World-Model-based Methods}} \\
            DrivingGPT\cite{chen2025drivinggpt}&98.9& 90.7& 94.9 &95.6 &79.7 &82.4\\
            Resim\cite{yang2026resim}&– &– &– &– &– &86.6\\
            DriveVLA-W0 \cite{li2025drivevla}&98.4& 95.3& 95.2 &100 &80.9& 87.2\\
            PWM\cite{georgiev2025pwm}& 98.6&  95.9&  95.4&  100 & 81.8&  88.1\\
            LAW~\cite{li2024enhancing_law}  & 96.4 & 95.4 & 88.7 & 99.9 & 81.7 & 84.6 \\
            World4Drive~\cite{zheng2025world4drive}  & 97.4 & 94.3 & 92.8 & 100 & 79.9 & 85.1 \\
            Epona~\cite{zhang2025epona}&   97.9&  95.1&  93.8&  99.9&  80.4&  86.2\\
            WoTE~\cite{wote} &  98.5 & 96.8 & 94.9 & 99.9 & 81.9 & 88.3 \\
            WorldRFT~\cite{yang2026worldrft} & 97.8 & 96.8 & 94.0 & 100 & 81.7 & 87.8 \\
            DriveLaW~\cite{xia2025drivelaw} &  99.0 & 97.1 & 96.7 & 100 & 81.3 & 89.1 \\
            DriveX-S\cite{shi2025drivex}&97.5&	94.0&	93.0&	100&	79.7&	84.5\\
            \cellcolor{red!5}$\operatorname{GraphWorld\ (Ours)}$ & \cellcolor{red!5}  99.0&\cellcolor{red!5} 97.1& \cellcolor{red!5}95.5& \cellcolor{red!5}100 &\cellcolor{red!5}83.2 &\cellcolor{red!5}90.1\\ 
            \bottomrule
    \end{tabular}}
\label{tab_NAVSIMv1}
\end{table}

\begin{table*}[t]
\centering
    \caption{Comparison with SOTA methods on the \textbf{NAVSIMv2 navhard} split~\cite{navsimv2}.}
\renewcommand\arraystretch{0.9}
  \tabcolsep=2.3mm %%%%%%%%%
  \resizebox{\linewidth}{!}{
\begin{tabular}{l c c ccccc ccccc}
\toprule
\multicolumn{1}{l}{Method}&Backbone&Stage&  NC$\uparrow$& DAC$\uparrow$& DDC$\uparrow$& TL$\uparrow$& EP$\uparrow$& TTC$\uparrow$& LK$\uparrow$& HC$\uparrow$& EC$\uparrow$& EPDMS$\uparrow$ \\
        \midrule
    \rowcolor{cyan!5}\multicolumn{13}{c}{\textit{E2E-based Methods}} \\
        \multirow{2}{*}{TransFuser~\cite{TransFuser}} 
     & \multirow{2}{*}{ResNet-34} & Stage 1 & 96.2 & 79.5 & 99.1 & 99.5 & 84.1 & 95.1 & 94.2 & 97.5 & 79.1 & \multirow{2}{*}{23.1} \\
     &  & Stage 2 & 77.7 & 70.2 & 84.2 & 98.0 & 85.1 & 75.6 & 45.4 & 95.7 & 75.9 &  \\
    \arrayrulecolor{gray!25}\cmidrule[0.01em]{1-13}\arrayrulecolor{black}
        \multirow{2}{*}{DiffusionDrive~\cite{liao2024diffusiondrive}} 
     & \multirow{2}{*}{ResNet-34} & Stage 1 & 96.0 & 79.7 & 97.4 & 99.5 & 81.3 & 93.1 & 90.8 & 96.8 & 73.8 & \multirow{2}{*}{24.2} \\
     &  & Stage 2 & 82.1 & 72.2 & 88.5 & 98.7 & 85.1 & 78.8 & 49.2 & 89.3 & 71.2 &  \\
    \arrayrulecolor{gray!25}\cmidrule[0.01em]{1-13}\arrayrulecolor{black}
        \multirow{2}{*}{GuideFlow~\cite{liu2025guideflow}} & \multirow{2}{*}{ResNet-34} & Stage 1 & 96.6 & 80.5 & 96.3 & 99.3 & 82.3 & 94.9 & 91.5 & 97.7 & 67.8 & \multirow{2}{*}{27.1} \\
         &  & Stage 2 & 87.3 & 76.7 & 88.8 & 99.2 & 84.3 & 85.1 & 49.7 & 93.1 & 44.5 &  \\
        \arrayrulecolor{gray!25}\cmidrule[0.01em]{1-13}\arrayrulecolor{black}
        \multirow{2}{*}{RAP-DINO} & \multirow{2}{*}{VIT} & Stage 1 
        & 97.1 & 94.4 & 98.8 & 99.8 & 83.9 & 96.9 & 94.7 & 96.4 & 66.2 & \multirow{2}{*}{36.9} \\
        &  & Stage 2 
        & 83.2 & 83.9 & 87.4 & 98.0 & 86.9 & 80.4 & 52.3 & 95.2 & 52.4 & \\

        \arrayrulecolor{gray!25}\cmidrule[0.01em]{1-13}\arrayrulecolor{black}
        \multirow{2}{*}{DriveSuprim~\cite{yao2025drivesuprim}} & \multirow{2}{*}{V2-99} & Stage 1 & 98.9 & 95.1 & 99.2 & 99.6 & 76.1 & 99.1 & 94.7 & 97.6 & 54.2 & \multirow{2}{*}{42.1} \\
         &  & Stage 2 & 87.9 & 88.8 & 89.6 & 98.8 & 80.3 & 86.0 & 53.5 & 97.1 & 56.1 &  \\

         % \multirow{2}{*}{GTRS-E~\cite{GTRS}} & \multirow{2}{*}{V2-99+EVA-ViT-L+ViT-L} & Stage 1 & 98.9 & 99.3 & 99.8 & 99.8 & 75.2 & 98.4 & 96.0 & 97.6 & 51.6 & \multirow{2}{*}{49.4} \\
         %  &  & Stage 2 & 92.3 & 93.3 & 94.6 & 99.2 & 73.1 & 91.2 & 53.9 & 96.7 & 56.8 &  \\
         %      \arrayrulecolor{gray!25}\cmidrule[0.01em]{1-13}\arrayrulecolor{black}
         %      \multirow{2}{*}{SimScale~\cite{tian2025simscale}} & \multirow{2}{*}{V2-99} & Stage 1 & 99.6 & 99.1 & 99.9 & 100.0 & 69.6 & 99.6 & 95.8 & 95.6 & 28.4 & \multirow{2}{*}{53.2} \\
         %       &  & Stage 2 & 94.5 & 94.2 & 95.8 & 99.2 & 75.8 & 92.8 & 60.1 & 96.1 & 43.2 &  \\
         %      \arrayrulecolor{gray!25}\cmidrule[0.01em]{1-13}\arrayrulecolor{black}
         %      \multirow{2}{*}{DrivoR~\cite{kirby2026driving}} & \multirow{2}{*}{ViT-S} & Stage 1 & 99.1 & 98.2 & 99.3 & 99.8 & 75.4 & 98.7 & 94.9 & 97.6 & 70.2 & \multirow{2}{*}{54.6} \\
         %       &  & Stage 2 & 92.3 & 91.6 & 97.3 & 99.1 & 75.7 & 90.6 & 56.1 & 98.4 & 44.7 &  \\
        \midrule
         \rowcolor{cyan!5}\multicolumn{13}{c}{\textit{VLA-based Methods}} \\
         \multirow{2}{*}{SpanVLA~\cite{zhou2026spanvla}} & \multirow{2}{*}{Qwen2.5-VL-3B} & Stage 1 & 98.4 & 94.3 & 97.8 & 99.9 & 85.7 & 97.2 & 94.2 & 97.6 & 72.1 & \multirow{2}{*}{40.1} \\
          &  & Stage 2 & 86.9 & 84.3 & 87.1 & 98.2 & 85.5 & 82.7 & 62.3 & 96.8 & 67.4 &  \\
         \arrayrulecolor{gray!25}\cmidrule[0.01em]{1-13}\arrayrulecolor{black}
         \multirow{2}{*}{DiffVLA~\cite{jiang2025diffvla}} & \multirow{2}{*}{V2-99 + ViT-L/14} & Stage 1 & 95.7 & 99.2 & 100.0 & 100.0 & 85.9 & 96.4 & 97.1 & 95.0 & 84.2 & \multirow{2}{*}{45.0} \\
          &  & Stage 2 & 81.2 & 88.8 & 94.6 & 99.0 & 86.0 & 76.4 & 59.8 & 98.6 & 80.4 &  \\

        % \multirow{2}{*}{HiST-VLA} & Stage 1 & 99.8 & 98.7 & 99.7 & 99.6 & 87.3 & 99.6 & 99.6 & 97.8 & 73.3 & \multirow{2}{*}{50.9} \\
        %  & Stage 2 & 86.4 & 83.7 & 89.9 & 98.5 & 87.5 & 83.1 & 56.7 & 96.3 & 60.9 &  \\
        \midrule
        \rowcolor{cyan!5}\multicolumn{13}{c}{\textit{World-Model-based Methods}} \\
        \multirow{2}{*}{MindDrive~\cite{suna2025minddrive}} & \multirow{2}{*}{ResNet-34} & Stage 1 & 96.1 & 86.0 & 98.8 & 99.3 & 83.3 & 95.6 & 94.4 & 97.6 & 74.7 & \multirow{2}{*}{30.9} \\
         &  & Stage 2 & 82.6 & 79.1 & 86.4 & 98.0 & 85.3 & 79.4 & 49.2 & 96.5 & 71.0 &  \\
        \arrayrulecolor{gray!25}\cmidrule[0.01em]{1-13}\arrayrulecolor{black}
        \multirow{2}{*}{World4Drive~\cite{zheng2025world4drive}} & \multirow{2}{*}{ResNet-34} & Stage 1 & 97.3 & 89.1 & 97.6 & 99.7 & 60.5 & 96.8 & 87.7 & 93.1 & 60.0 & \multirow{2}{*}{34.9} \\
         &  & Stage 2 & 91.4 & 82.0 & 91.0 & 98.5 & 53.1 & 90.6 & 52.3 & 93.3 & 62.8 &  \\
        \arrayrulecolor{gray!25}\cmidrule[0.01em]{1-13}\arrayrulecolor{black}

        \rowcolor{red!5}  & & Stage 1 & 98.1 & 97.9 & 97.9 & 98.0 & 83.9 & 97.8
        & 97.2 & 95.4 & 64.5 &  \\
        \rowcolor{red!5} \multirow{-2}{*}{GraphWorld (Ours)}& \multirow{-2}{*}{ResNet-34} & Stage 2 & 88.6 & 85.9 & 92.2 & 97.6 & 82.4 & 87.0 & 56.6 & 91.4 & 44.0 &  \multirow{-2}{*}{$\pmb{53.6}$}\\
\bottomrule
\end{tabular}}
\label{tab_navsimv2_navhard}
\end{table*}

% \begin{table}[!htbp]
\begin{table*}[t]
\centering
    \caption{Comparison with SOTA methods on the \textbf{NAVSIMv2 navtest} split~\cite{navsimv2}. EPDMS$^{*}$ denotes results computed with the original NAVSIMv2 evaluation code before the human-behavior filtering fix, while EPDMS denotes the corrected official implementation.}
\renewcommand\arraystretch{0.9}
  \tabcolsep=3.3mm %%%%%%%%%
  \resizebox{\linewidth}{!}{
\begin{tabular}{l c ccccc cccc}
\toprule
\multicolumn{1}{l}{Method}&  NC$\uparrow$& DAC$\uparrow$& DDC$\uparrow$& TL$\uparrow$& EP$\uparrow$& TTC$\uparrow$& LK$\uparrow$& HC$\uparrow$& EC$\uparrow$& EPDMS$\uparrow$ \\
        \midrule
        \rowcolor{cyan!5}\multicolumn{11}{c}{\textit{E2E-based Methods}} \\
        TransFuser~\cite{TransFuser} & 96.9 & 89.9 & 97.8 & 99.7 & 87.1 & 95.4 & 92.7 & 98.3 & 87.2 & 76.7 \\
        DiffusionDrive~\cite{liao2024diffusiondrive} & 98.2 & 95.9 & 99.4 & 99.8 & 87.5 & 97.3 & 96.8 & 98.3 & 87.7  &84.5 \\
        Hydra-MDP++~\cite{li2025hydraplus} & 97.2 & 97.5 & 99.4 & 99.6 & 83.1 & 96.5 & 94.4 & 98.2 & 70.9 & 81.4  \\
        DriveSuprim~\cite{yao2025drivesuprim} & 97.5 & 96.5 & 99.4 & 99.6 & 88.4 & 96.6 & 95.5 & 98.3 & 77.0 & 83.1  \\
        DiffusionDriveV2~\cite{zou2025diffusiondrivev2} & 97.7 & 96.6 & 99.2 & 99.8 & 88.9 & 97.2 & 96.0 & 97.8 & 91.0 &  87.5 \\
        \midrule
        \rowcolor{cyan!5}\multicolumn{11}{c}{\textit{VLA-based Methods}} \\
        DriveWorld-VLA~\cite{liu2026driveworld} & 98.6 & 99.1 & 99.6 & 99.8 & 87.4 & 97.9 & 97.0 & 97.8 & 78.6 & 86.8 \\
        DriveVLA-W0~\cite{li2025drivevla} & 98.5 & 99.1 & 98.0 & 99.7 & 86.4 & 98.1 & 93.2 & 97.9 & 58.9 &  86.1 \\
        Recogdrive~\cite{li2025recogdrive} & 98.3 & 95.2 & 98.3 & 99.8 & 87.1 & 97.5 & 96.6 & 99.5 & 86.5 &  83.6 \\
        \midrule
        \rowcolor{cyan!5}\multicolumn{11}{c}{\textit{World-Model-based Methods}} \\
        Latent-WAM~\cite{wang2026latentwam} & 98.1 & 97.3 & 99.6 & 99.8 & 87.7 & 97.3 & 97.6 & 98.1 & 87.3 &  89.3 \\
        \rowcolor{red!5} GraphWorld (Ours) & 98.4 &98.8 &99.1& 99.1& 85.9 &97.9& 96.0& 97.8 &74.6&  89.5 \\
        \bottomrule
\end{tabular}}
\label{tab_navsimv2_navtest}
\end{table*}

\begin{table}[t]
\scriptsize
\centering
\caption[]{Planning results for \textbf{3-second short-horizon planning} on the \textbf{nuScenes} validation set.}
\renewcommand\arraystretch{0.7}
\setlength{\tabcolsep}{0.99mm}
  \resizebox{\linewidth}{!}{
\begin{tabular}{l cccc  cccc }
\toprule
\multirow{2}{*}{$\operatorname{Method}$} & \multicolumn{4}{c}{$\operatorname{L2\ (m)}\downarrow$} & \multicolumn{4}{c}{$\operatorname{Col.\ Rate\ (\%)}\downarrow$}  \\
\cmidrule(lr){2-5} \cmidrule(lr){6-9}
& 1s & 2s & 3s &  $\operatorname{Avg.}$ & 1s & 2s & 3s &$\operatorname{Avg.}$ \\
\midrule
$\operatorname{UniAD}$~\cite{uniad} &0.48& 0.96& 1.65& \cellcolor{red!5}1.03& 0.10& 0.15& 0.61& \cellcolor{red!5}0.29 \\
$\operatorname{VAD}$~\cite{jiang2023vad} &0.41& 0.70& 1.05& \cellcolor{red!5}0.72 &0.11& 0.24& 0.42 &\cellcolor{red!5}0.26 \\
$\operatorname{DiffusionDrive}$~\cite{liao2024diffusiondrive} &0.29& 0.58& 0.96& \cellcolor{red!5}0.61& 0.02& 0.05& 0.22& \cellcolor{red!5}0.09 \\
$\operatorname{DIVER}$~\cite{diver} &-& -& -&\cellcolor{red!5} -& 0.01& 0.05& \textbf{0.15}& \cellcolor{red!5}\textbf{0.07} \\
$\operatorname{FocalAD}$~\cite{sun2025focalad} &0.27& 0.57& 0.96& \cellcolor{red!5}0.60 &\textbf{0.00}& \textbf{0.04} &0.24& \cellcolor{red!5}0.09 \\
% $\operatorname{GuideFlow}$~\cite{liu2025guideflow} &-& -& -& \cellcolor{red!5}-&0.00& 0.02& 0.18& \cellcolor{red!5}0.07 \\
$\operatorname{SparseDrive}$~\cite{sun2024sparsedrive} &0.30& 0.58&0.96& \cellcolor{red!5}0.61& 0.01 &0.05& 0.23&\cellcolor{red!5} 0.10 \\
$\operatorname{LAW}$~\cite{li2024enhancing_law}&0.26 &0.57 &1.01 &\cellcolor{red!5}0.61& 0.14 &0.21& 0.54 &\cellcolor{red!5}0.30\\
$\operatorname{GenAD}$~\cite{zheng2024genad}&0.28 &0.49 &0.78 &\cellcolor{red!5}0.52 &0.08 &0.14& 0.34& \cellcolor{red!5}0.19\\
$\operatorname{Drive-OccWorld}$~\cite{zheng2024occworld}&0.25& 0.44 &0.72& \cellcolor{red!5}0.47 &0.03 &0.08 &0.22 &\cellcolor{red!5}0.11\\
$\operatorname{SSR}$~\cite{ssr}&\textbf{0.19}& \textbf{0.36}& \textbf{0.62} &\cellcolor{red!5}\textbf{0.39} &0.10 &0.10& 0.24& \cellcolor{red!5}0.15\\
$\operatorname{MomAD}$~\cite{momad}&0.31& 0.57 &0.91 &\cellcolor{red!5}0.60& 0.01& 0.05& 0.22& \cellcolor{red!5}0.09\\
\cellcolor{red!5}$\operatorname{GraphWorld (Ours)}$&\cellcolor{red!5}0.29& \cellcolor{red!5}0.55 &\cellcolor{red!5}0.89 &\cellcolor{red!5}0.57& \cellcolor{red!5}\textbf{0.00}&\cellcolor{red!5} \textbf{0.04}& \cellcolor{red!5}0.20&\cellcolor{red!5}0.08\\
\bottomrule
\end{tabular}}
\label{tab_nuscenes_planning_3s}
\end{table}

\begin{table}[t]
\centering
  \caption{\textbf{Motion prediction} results on the nuScenes dataset.}
\renewcommand\arraystretch{0.7}
  \tabcolsep=2.6mm %%%%%%%%%
  \resizebox{\linewidth}{!}{
\begin{tabular}{l cccc}
\toprule
\multicolumn{1}{l}{Method}& minADE$\downarrow$                & minFDE$\downarrow$ & MR$\downarrow$                 & EPA$\uparrow$    \\
\midrule
$\operatorname{UniAD}$  \cite{uniad}& \cellcolor{red!5}0.71& 1.02 &0.151& 0.456 \\
$\operatorname{VAD}$  \cite{jiang2023vad}& \cellcolor{red!5}0.68& 0.88 &\textbf{0.083}& - \\
$\operatorname{SparseDrive}$  \cite{sun2024sparsedrive}& \cellcolor{red!5}0.62& 0.99 &0.136& {0.482} \\
$\operatorname{GraphWorld (Ours)}$& \cellcolor{red!5}\textbf{0.55}& \textbf{0.86} &0.112& \textbf{0.512} \\
\bottomrule
\end{tabular}}
\label{tab_nuscenes_motion}
\end{table}

\begin{table}[t]
\small
\centering
  \caption{Robustness study of planning performance on \textbf{Adv-nuSc} under adversarial driving scenarios.}
  \renewcommand\arraystretch{0.7}
  \setlength{\tabcolsep}{4.0mm}% Adjust column spacing
  \resizebox{\linewidth}{!}{
  \begin{tabular}{l cccc}
\toprule
\multirow{2}{*}{$\operatorname{Method}$}& \multicolumn{4}{c}{$\operatorname{Col.\ Rate\ (\%)}\downarrow$} \\
\cmidrule(lr){2-5}
& 1s & 2s & 3s & $\operatorname{Avg.}$  \\
\midrule
$\operatorname{UniAD}$ \cite{uniad}&0.800   & 4.100 & 6.960 & \cellcolor{red!5}3.950 \\
$\operatorname{VAD}$ \cite{jiang2023vad} &4.460 &7.590 &9.080 &\cellcolor{red!5}7.050 \\
$\operatorname{SparseDrive}$ \cite{sun2024sparsedrive}&0.029& 0.618 &2.430&\cellcolor{red!5}1.026 \\
$\operatorname{DiffusionDrive}$ \cite{liao2024diffusiondrive}&0.068& 1.299&3.646 &\cellcolor{red!5}1.671 \\

$\operatorname{DIVER}$ \cite{diver} &0.033 & 0.423 & 1.798     & \cellcolor{red!5}0.752\\

\cellcolor{red!5}$\operatorname{GraphWorld\ (Ours)}$ &\cellcolor{red!5}\textbf{0.028} & \cellcolor{red!5}\textbf{0.420} &    \cellcolor{red!5}\textbf{1.780}      & \cellcolor{red!5}\textbf{0.742}\\
\bottomrule
\end{tabular}}
\label{tab_advnusc_planning}
\end{table}

\begin{table}[t]
\small
\centering
  \caption{Robustness study of planning performance on \textbf{nuScenes-C}.}
  \renewcommand\arraystretch{0.7}
  \setlength{\tabcolsep}{3.0mm}% Adjust column spacing
  \resizebox{\linewidth}{!}{
  \begin{tabular}{ll cccc}
\toprule
\multirow{2}{*}{$\operatorname{Scene}$}&\multirow{2}{*}{$\operatorname{Method}$}& \multicolumn{4}{c}{$\operatorname{Col.\ Rate\ (\%)}\downarrow$} \\
\cmidrule(lr){3-6}
&& 1s & 2s & 3s & $\operatorname{Avg.}$  \\

\midrule
\multirow{4}{*}{$\operatorname{Clean}$}&$\operatorname{SparseDrive}$ \cite{sun2024sparsedrive}& 0.01& 0.05& 0.18 & \cellcolor{red!5}0.08 \\
&$\operatorname{MomAD}$ \cite{momad}& 0.01& 0.05& 0.22 & \cellcolor{red!5}0.09 \\
&$\operatorname{DiffusionDrive}$ \cite{liao2024diffusiondrive}& 0.03& 0.05& \textbf{0.16}& \cellcolor{red!5}0.08 \\
&\cellcolor{red!5}$\operatorname{GraphWorld\ (Ours)}$ &\cellcolor{red!5}\textbf{0.00}& \cellcolor{red!5}\textbf{0.04} &\cellcolor{red!5}0.20 &\cellcolor{red!5}\textbf{0.08}\\

\midrule
\multirow{4}{*}{$\operatorname{Snow}$}&$\operatorname{SparseDrive}$ \cite{sun2024sparsedrive}& 0.13&0.27&0.50 & \cellcolor{red!5}0.30 \\
&$\operatorname{MomAD}$ \cite{momad}& 0.08&0.16&0.30 & \cellcolor{red!5}0.18 \\
&$\operatorname{DiffusionDrive}$ \cite{liao2024diffusiondrive}& 0.09& 0.24& 0.39& \cellcolor{red!5}0.24 \\
&\cellcolor{red!5}$\operatorname{GraphWorld\ (Ours)}$ &\cellcolor{red!5}\textbf{0.07}&\cellcolor{red!5}\textbf{0.15}&\cellcolor{red!5}\textbf{0.28}& \cellcolor{red!5}\textbf{0.17}\\

\midrule
\multirow{4}{*}{$\operatorname{Rain}$}&$\operatorname{SparseDrive}$ \cite{sun2024sparsedrive}& 0.11& 0.27& 0.55 & \cellcolor{red!5}0.31 \\
&$\operatorname{MomAD}$ \cite{momad}&0.06 &0.17 & 0.31 & \cellcolor{red!5}0.18 \\
&$\operatorname{DiffusionDrive}$ \cite{liao2024diffusiondrive}& 0.07&0.18& 0.35& \cellcolor{red!5}0.20 \\
&\cellcolor{red!5}$\operatorname{GraphWorld\ (Ours)}$ &\cellcolor{red!5}\textbf{0.05}& \cellcolor{red!5}\textbf{0.15}& \cellcolor{red!5}\textbf{0.29}& \cellcolor{red!5}\textbf{0.16}\\

\midrule
\multirow{4}{*}{$\operatorname{Fog}$}&$\operatorname{SparseDrive}$ \cite{sun2024sparsedrive}&0.14 &0.36& 0.58 & \cellcolor{red!5}0.36 \\
&$\operatorname{MomAD}$ \cite{momad}& 0.06&0.19& 0.32& \cellcolor{red!5}0.19 \\
&$\operatorname{DiffusionDrive}$ \cite{liao2024diffusiondrive}& 0.06&0.18& 0.30& \cellcolor{red!5}0.18 \\
&\cellcolor{red!5}$\operatorname{GraphWorld\ (Ours)}$ & \cellcolor{red!5}\textbf{0.05}& \cellcolor{red!5}\textbf{0.17}& \cellcolor{red!5}\textbf{0.29}& \cellcolor{red!5}\textbf{0.17}\\
\bottomrule
\end{tabular}}
\label{tab_nuScenes_C_appendix}
\end{table}

\begin{table}[t]
\small
\centering
  \caption{Robustness study of planning results on the  \textbf{Turning\text{-}nuScenes} \cite{momad} validation dataset. $^*$ denotes the re-implementation.}
  \renewcommand\arraystretch{0.7}
  \setlength{\tabcolsep}{5.0mm}% Adjust column spacing
  \resizebox{\linewidth}{!}{
  \begin{tabular}{l cccc}
\toprule
\multirow{2}{*}{$\operatorname{Method}$}&  \multicolumn{4}{c}{$\operatorname{Col.\ Rate\ (\%)}\downarrow$} \\
\cmidrule(lr){2-5}
& 1s & 2s & 3s & $\operatorname{Avg.}$   \\
\midrule
$\operatorname{SparseDrive}$ \cite{sun2024sparsedrive}
& 0.04 & 0.17 & 0.98 & \cellcolor{red!5}0.40  \\
$\operatorname{DiffusionDrive}^{*}$ \cite{liao2024diffusiondrive} & 
\textbf{0.03}& 
0.14 & 
0.85 & 
\cellcolor{red!5}0.34 \\
$\operatorname{MomAD}$ \cite{momad} & 
\textbf{0.03}& 
0.13 & 
0.79 & 
\cellcolor{red!5}0.32 \\
\cellcolor{red!5}$\operatorname{GraphWorld\ (Ours)}$& 
\cellcolor{red!5}\textbf{0.03}& 
\cellcolor{red!5}\textbf{0.12}& 
\cellcolor{red!5}\textbf{0.72}& 
\cellcolor{red!5}\textbf{0.28}\\
\bottomrule
\end{tabular}}
\label{tab_turninguscenes_planning_appendix}
\end{table}

\begin{table*}[!htp]
\centering
\caption{Ablation study of different modules across \textbf{nuScenes} (open-loop), \textbf{NAVSIMv1} (closed-loop), and \textbf{Bench2Drive}.}
\renewcommand\arraystretch{0.75}
\setlength{\tabcolsep}{1.2mm}
\resizebox{\linewidth}{!}{
\begin{tabular}{cc|cccccc|cccccc|cccc}
\toprule
& & \multicolumn{6}{c|}{\textbf{nuScenes (Open-loop)}} 
& \multicolumn{6}{c|}{\textbf{NAVSIMv1 (Closed-loop)}} 
& \multicolumn{4}{c}{\textbf{Bench2Drive}} \\

\cmidrule(lr){3-8} \cmidrule(lr){9-14} \cmidrule(lr){15-18}

ECIG & WSCP 
& L2@4s$\downarrow$ & L2@5s$\downarrow$ & L2@6s$\downarrow$
& Col.@4s$\downarrow$ & Col.@5s$\downarrow$ & Col.@6s$\downarrow$
& NC$\uparrow$ & DAC$\uparrow$ & TTC$\uparrow$ & Comf.$\uparrow$ & EP$\uparrow$ & PDMS$\uparrow$
& DS$\uparrow$ & SR$\uparrow$ & Effi$\uparrow$ & Comf$\uparrow$ \\

\midrule

& 
& 1.75 & 2.32 & 2.95
& 0.87 & 1.54 & 2.33
& 96.2 & 95.4 & 90.7 & 100 & 80.7 & 85.1
& 44.54 & 16.71 & 170.21 & 48.63 \\

$\checkmark$ & 
& 1.71 & 2.28 & 2.76
& 0.79 & 1.47 & 2.19
& 97.2 & 96.0 & 91.6 & 100 & 81.1 & 85.5
& 49.44 & 23.89 & 177.92 & 53.07 \\

\rowcolor{red!5}
$\checkmark$ & $\checkmark$
& \textbf{1.64} & \textbf{1.88} & \textbf{2.29}
& \textbf{0.65} & \textbf{1.29} & \textbf{1.95}
& \textbf{99.0} & \textbf{97.1} & \textbf{95.5} & \textbf{100} & \textbf{83.2} & \textbf{90.1}
& \textbf{51.55} & \textbf{25.47} & \textbf{181.12} & \textbf{56.59} \\

\bottomrule
\end{tabular}}
\label{tab_ablation_modules_navsim_nuscenes_b2d}
\end{table*}

% \begin{table}[t]
% \centering
% \caption{Ablation study of different modules on the \textbf{NAVSIM} dataset (closed-loop).
% We use ${\operatorname{TransFuser}_{\operatorname{mmt}}}^{*}$ as the baseline.}
% \renewcommand\arraystretch{0.7}
% \setlength{\tabcolsep}{2.6mm}
% \resizebox{\linewidth}{!}{
% \begin{tabular}{cc cccccc}
% \toprule
% ECIG & WSCP 
% & NC$\uparrow$ & DAC$\uparrow$ & TTC$\uparrow$ & Comf.$\uparrow$ & EP$\uparrow$ & PDMS$\uparrow$ \\
% \midrule
% & 
% & 96.2 & 95.4 & 90.7 & 100 & 80.7 & 85.1 \\
% $\checkmark$ & 
% & 97.2 & 96.0 & 91.6 & 100 & 81.1 & 85.5 \\
% \cellcolor{red!5}$\checkmark$ & \cellcolor{red!5}$\checkmark$
% & \cellcolor{red!5}99.0 & \cellcolor{red!5}97.1 & \cellcolor{red!5}95.5
% & \cellcolor{red!5}100 & \cellcolor{red!5}83.2 & \cellcolor{red!5}89.1 \\
% \bottomrule
% \end{tabular}}
% \label{tab_ablation_navsim_modules}
% \end{table}

\begin{table*}[t]
\centering
\caption{Ablation study on different graph structures across \textbf{NAVSIMv1} and \textbf{nuScenes}. 
\emph{Fully-connected} denotes a dense graph, while \emph{ego→nbrs} denotes an ego-centric star graph.}
\renewcommand\arraystretch{0.75}
\setlength{\tabcolsep}{2.5mm}
\resizebox{\linewidth}{!}{
\begin{tabular}{l|cccccc|cccccc}
\toprule
& \multicolumn{6}{c|}{\textbf{NAVSIMv1}} 
& \multicolumn{6}{c}{\textbf{nuScenes}} \\
\cmidrule(lr){2-7} \cmidrule(lr){8-13}

Structure 
& NC$\uparrow$ & DAC$\uparrow$ & TTC$\uparrow$ & Comf.$\uparrow$ & EP$\uparrow$ & PDMS$\uparrow$
& L2@4s$\downarrow$ & L2@5s$\downarrow$ & L2@6s$\downarrow$
& Col.@4s$\downarrow$ & Col.@5s$\downarrow$ & Col.@6s$\downarrow$ \\

\midrule

Fully-connected 
& 96.1 & 95.2 & 90.3 & 100 & 80.6 & 85.0
& 1.73 & 1.98 & 2.45 & 0.68 & 1.40 & 2.23 \\

\rowcolor{red!5}
ego→nbrs 
& \textbf{99.0} & \textbf{97.1} & \textbf{95.5} & \textbf{100} & \textbf{83.2} & \textbf{90.1}
& \textbf{1.64} & \textbf{1.88} & \textbf{2.29} & \textbf{0.65} & \textbf{1.29} & \textbf{1.95} \\

\bottomrule
\end{tabular}}
\label{tab_ablation_graph_structure_navsimv1_nuscenes}
\end{table*}

\begin{table*}[t]
\scriptsize
\centering
\caption{Comparison of Diffusion and Flow-Matching across different datasets.}
\renewcommand\arraystretch{0.8}
\setlength{\tabcolsep}{2.5mm}
\resizebox{\linewidth}{!}{
\begin{tabular}{l|ccc|ccc|cccc|cccccc}
\toprule
\multirow{2}{*}{Strategies} 
& \multicolumn{6}{c|}{\textbf{nuScenes}} 
& \multicolumn{4}{c|}{\textbf{Bench2Drive}} 
& \multicolumn{6}{c}{\textbf{NAVSIMv1}} \\
\cmidrule(lr){2-7} \cmidrule(lr){8-11} \cmidrule(lr){12-17}

& \multicolumn{3}{c|}{L2 (m)$\downarrow$} 
& \multicolumn{3}{c|}{Col. Rate (\%)$\downarrow$}
& DS$\uparrow$ & SR$\uparrow$ & Effi$\uparrow$ & Comf$\uparrow$
& NC$\uparrow$ & DAC$\uparrow$ & TTC$\uparrow$ & Comf$\uparrow$ & EP$\uparrow$ & PDMS$\uparrow$ \\

\cmidrule(lr){2-4} \cmidrule(lr){5-7}
& 4s & 5s & 6s & 4s & 5s & 6s 
& & & & 
& & & & & & \\
\midrule

Diffusion 
& 1.79 & 2.07 & 2.46 & 0.75 & 1.45 & 2.23
& 73.22 & 40.74 & 77.92 & 25.81
& 98.0 & 96.6 & 94.8 & 100 & 82.6 & 88.1 \\

\rowcolor{red!5}
Flow-Matching 
& \textbf{1.64} & \textbf{1.88} & \textbf{2.29} 
& \textbf{0.65} & \textbf{1.29} & \textbf{1.95}
& \textbf{76.71} & \textbf{44.22} & \textbf{80.48} & \textbf{29.74}
& \textbf{99.0} & \textbf{97.1} & \textbf{95.5} & \textbf{100} & \textbf{83.2} & \textbf{90.1} \\

\bottomrule
\end{tabular}}
\label{tab_all_datasets_diffusion_vs_flow}
\end{table*}

\begin{table}[t]
\scriptsize
\centering
\caption[]{Ablation study of different stages on the \textbf{nuScenes} set.}
\renewcommand\arraystretch{0.7}
\setlength{\tabcolsep}{3.39mm}
  \resizebox{\linewidth}{!}{
\begin{tabular}{l ccc  ccc }
\toprule
\multirow{2}{*}{$\operatorname{Stage}$} & \multicolumn{3}{c}{$\operatorname{L2\ (m)}\downarrow$} & \multicolumn{3}{c}{$\operatorname{Col.\ Rate\ (\%)}\downarrow$}  \\
\cmidrule(lr){2-4} \cmidrule(lr){5-7}
&  4s & 5s & 6s&   4s & 5s & 6s \\
\midrule
One&1.68 &1.96& 2.40 & 0.68 &1.35 &2.04\\
\cellcolor{red!5}Two&\cellcolor{red!5}\textbf{1.64} &\cellcolor{red!5}\textbf{1.88}& \cellcolor{red!5}\textbf{2.29} &\cellcolor{red!5}\textbf{0.65} &\cellcolor{red!5}\textbf{1.29} &\cellcolor{red!5}\textbf{1.95}\\

\bottomrule
\end{tabular}}
\label{tab_nuscenes_planning_6s_different_stages}
\end{table}

\begin{table}[t]
\scriptsize
\centering
\caption[]{Ablation study of \textbf{neighbor selection strategies} on the \textbf{nuScenes} validation set.
The upper rows use distance-based selection, while the lower rows adopt Top-$K$ neighbors. We use ${\operatorname{SparseDrive}}$ as the baseline.}
\renewcommand\arraystretch{0.7}
\setlength{\tabcolsep}{3.0mm}
  \resizebox{\linewidth}{!}{
\begin{tabular}{c ccc  ccc }
\toprule
\multirow{2}{*}{$\operatorname{H-Param}$}  & \multicolumn{3}{c}{$\operatorname{L2\ (m)}\downarrow$} & \multicolumn{3}{c}{$\operatorname{Col.\ Rate\ (\%)}\downarrow$}  \\
\cmidrule(lr){2-4} \cmidrule(lr){5-7}
&  4s & 5s & 6s &  4s & 5s & 6s  \\
\midrule
 5m  &1.70 &1.95& 2.35& 0.72 &1.36 &2.02 \\
 8m  &1.68 &1.91 &2.33 &0.68 &1.33 &1.98 \\
 \cellcolor{red!5}10m  &\cellcolor{red!5}\textbf{1.64} &\cellcolor{red!5}\textbf{1.88}&\cellcolor{red!5}\textbf{2.29}&\cellcolor{red!5}\textbf{0.65} &\cellcolor{red!5}\textbf{1.29} &\cellcolor{red!5}\textbf{1.95} \\
 12m  &1.65 &1.90 &2.31 &0.67 &1.32 &1.96 \\
 15m  &1.66 &1.91 &2.32 &0.68 &1.34 &1.99 \\
 \midrule
 4  &1.71 &1.93 &2.35 &0.70 &1.34 &2.01 \\
 8  &1.66 &1.90& 2.32& 0.68 &1.31 &1.99 \\
 16  &1.73& 2.02 &2.43&  0.75 &1.44& 2.09 \\
 32  &1.75& 2.22 &2.72&  0.82 &1.50& 2.23 \\
\bottomrule
\end{tabular}}
\label{tab_Ablation_nuscenes_neighbor_selection_strategies}
\end{table}

\begin{table}[t]
\scriptsize
\centering
\caption[]{Ablation study of different times on the \textbf{nuScenes} set.}
\renewcommand\arraystretch{0.7}
\setlength{\tabcolsep}{3.39mm}
  \resizebox{\linewidth}{!}{
\begin{tabular}{l ccc  ccc }
\toprule
\multirow{2}{*}{$\operatorname{Stage}$} & \multicolumn{3}{c}{$\operatorname{L2\ (m)}\downarrow$} & \multicolumn{3}{c}{$\operatorname{Col.\ Rate\ (\%)}\downarrow$}  \\
\cmidrule(lr){2-4} \cmidrule(lr){5-7}
&  4s & 5s & 6s&   4s & 5s & 6s \\
\midrule
\cellcolor{red!5}$t+1$&\cellcolor{red!5}\textbf{1.64} &\cellcolor{red!5}\textbf{1.88}&\cellcolor{red!5}\textbf{2.29} & \cellcolor{red!5}\textbf{0.65} &\cellcolor{red!5}\textbf{1.29} &\cellcolor{red!5}\textbf{1.95}\\
$t+2$&1.75 &1.99& 2.44 & 0.69 &1.38 &2.09\\
$t+3$&1.81 &2.06& 2.57 & 0.74 &1.47 &2.26\\
\bottomrule
\end{tabular}}
\label{tab_nuscenes_planning_6s_different_times_appendix}
\end{table}

\begin{table}[t]
\scriptsize
\centering
\caption{Ablation on sampling steps on \textbf{NAVSIM}.}
\renewcommand\arraystretch{0.7}
\setlength{\tabcolsep}{2.0mm}
\resizebox{\linewidth}{!}{
\begin{tabular}{c cccccc c}
\toprule
Steps & NC$\uparrow$ & DAC$\uparrow$ & TTC$\uparrow$ & Comf.$\uparrow$ & EP$\uparrow$ & PDMS$\uparrow$ & FPS$\uparrow$ \\
\midrule
1 & 98.6 & 96.7 & 94.9 & 100 & 82.4 & 88.3 & 51 \\
\rowcolor{red!5}
2 & \textbf{99.0} & \textbf{97.1} & \textbf{95.5} & \textbf{100} & \textbf{83.2} & \textbf{90.1} & 51 \\
3 & 99.1 & 97.2 & 95.6 & 100 & 83.3 & 89.2 & 46 \\
4 & 99.1 & 97.2 & 95.7 & 100 & 83.3 & 89.3 & 42 \\
\bottomrule
\end{tabular}}
\label{tab_sampling_steps}
\end{table}

\begin{table}[!htp]
\scriptsize
\centering
\caption{Ablation on solver choice on \textbf{NAVSIM}.}
\renewcommand\arraystretch{0.7}
\setlength{\tabcolsep}{2.0mm}
\resizebox{\linewidth}{!}{
\begin{tabular}{l cccccc c}
\toprule
Solver & NC$\uparrow$ & DAC$\uparrow$ & TTC$\uparrow$ & Comf.$\uparrow$ & EP$\uparrow$ & PDMS$\uparrow$ & FPS$\uparrow$ \\
\midrule
\rowcolor{red!5}Euler & 99.0 & 97.1 & 95.5 & 100 & 83.2 & 90.1 & 51 \\
Heun  & 99.1 & 97.2 & 95.6 & 100 & 83.3 & 89.2 & 46 \\
\bottomrule
\end{tabular}}
\label{tab_solver_choice}
\end{table}

\section{Experiments}

\subsection{Dataset}

\noindent \textbf{Bench2Drive  (Close-Loop).}
We conduct training and evaluation of GraphWorld on the \textbf{Bench2Drive} \cite{jia2024bench2drive}, a closed-loop evaluation protocol based on the CARLA Leaderboard 2.0 \cite{CARLA} for E2E-AD. It provides a base training set of 1000 clips, with 950 used for training and 50 for open-loop validation. Each clip captures approximately 150 meters of continuous driving in a specific traffic scenario. For closed-loop evaluation, we use the official 220 routes, covering 44 interactive scenarios with 5 routes each.

\noindent \textbf{NAVSIMv1/2  (Close-Loop).}
We conduct training and evaluation of GraphWorld on the \textbf{NAVSIMv1} \cite{navsimv1} dataset. NAVSIMv1 is a real-world, planning-oriented dataset that builds upon OpenScene, a compact redistribution of nuPlan \cite{caesar2021nuplan}, the largest publicly available annotated driving dataset. It leverages eight cameras to achieve a full 360° field of view, along with a merged LiDAR point cloud derived from five sensors. Annotations are provided at 2 Hz and include both HD maps and object bounding boxes. The dataset is specifically designed to emphasize challenging driving scenarios involving dynamic changes in driving intentions, while deliberately excluding trivial cases such as stationary scenes or constant-speed cruising.

\noindent \textbf{NuScenes (Open-Loop).}
We conduct extensive open-loop experiments on the \textbf{nuScenes} dataset \cite{nuscenes}, which consists of 1000 driving scenes (700 for training, 150 for validation and 150 for test). Each scene lasts 20 seconds and includes around 40 key-frames annotated at 2 Hz. Each sample contains six images from surround-view cameras (covering 360° FOV), and point clouds from both LiDAR and radar sensors. 

\noindent \textbf{Turning-nuScenes (Open-Loop).}
We conduct extensive open-loop experiments on the \textbf{Turning-nuScenes} dataset \cite{momad}, a challenging subset of NuScenes proposed by MomAD \cite{momad} to evaluate trajectory consistency in non-trivial maneuvers. While most planning tasks in the original nuScenes dataset primarily involve go-straight commands, Turning-nuScenes specifically focuses on turning scenarios to assess the temporal coherence of predicted trajectories. To construct this subset, samples are selected based on a displacement threshold of 25 meters between the predicted positions at 0.5s and 3.0s in the GT ego trajectory. The resulting validation set comprises 680 samples across 17 scenes, accounting for approximately one-tenth of the full nuScenes validation set.

\noindent \textbf{Adv-nuSc  (Open-Loop).}
To evaluate adversarial robustness, we conduct extensive open-loop experiments on the \textbf{Adv-nuSc} \cite{xu2025challenger} dataset. It contains 156 scenes (6,115 samples) and is specifically crafted to challenge the ego vehicle by introducing adversarial traffic participants.It is built upon the validation split of the nuScenes dataset  \cite{nuscenes}, which contains 150 scenes, each with 20 seconds of driving data. For each scene, we randomly select up to 10 background vehicles (if there are that many) that come close to the ego vehicle at any point in time and designate them as candidate adversarial agents. Challenger is then used to generate adversarial trajectories for these vehicles, creating diverse and challenging driving scenarios.

\noindent \textbf{NuScenes-C  (Open-Loop).}
\textbf{NuScenes-C} \cite{zhujun_benchmarking} is a corrupted benchmark derived from the nuScenes validation set, introducing various types of noise to assess the robustness of planning models. It includes 27 corruption types applied at 5 severity levels. To evaluate robustness under adverse weather conditions, we select three representative weather corruptions — Rain, Snow, and Fog — as our test scenarios.

\subsection{Evaluation Metrics}
\noindent\textbf{Close-Loop (Bench2Drive).} The Bench2Drive \cite{jia2024bench2drive} includes five metrics for closed-loop evaluation: Driving Score (DS), Success Rate (SR), Efficiency, Comfortness, and Multi-Ability. The Success Rate quantifies the proportion of routes successfully completed within the allotted time. The Driving Score follows CARLA [11], incorporating both route completion status and violation penalties, where infractions reduce the score via discount factors. Efficiency and Comfortness are used to measure the speed performance and comfort of the autonomous driving system during the driving process, respectively. Multi-Ability measures 5 advanced skills, including `Merging, Overtaking, Emergency Brake, Give Way, and Traffic Sign', independently for urban driving. 

\noindent\textbf{Close-Loop (NAVSIMv1).}
NAVSIMv1\cite{navsimv1} metrics include No at-fault Collision (NC), Drivable Area Compliance (DAC), Time-to-Collision (TTC), Comfort (C.), and Ego Progress (EP). NAVSIMv1 uses the Predictive Driver Model Score (PDMS) to evaluate model performance.

% \noindent\textbf{Close-Loop (NAVSIMv2).}
% NAVSIMv2\cite{navsimv2}includes several components, categorized as penalties or weighted subscores. Key metrics are No at-fault Collision (NC), Drivable Area Compliance (DAC), Driving Direction Compliance (DDC), Traffic Light Compliance (TLC), Ego Progress (EP), Time to Collision (TTC), Lane Keeping (LK), History Comfort (HC), and Extended Comfort (EC). NAVSIMv2 uses the Extended Predictive Driver Model Score (EPDMS) to evaluate model performance. Furthermore, NAVSIMv2 introduces an additional Navhard split, which adopts a twostage evaluation workflow. In Stage I, the planner
% predicts a trajectory from the real observation and receives an initial score EPDMS. In Stage II, multiple plausible future observations are synthesized around
% the Stage 1 endpoint using 3D Gaussian Splatting, and
% the planner is re-evaluated under these perturbed conditions to obtain EPDMS. The two scores are fused
% through Gaussian-weighted aggregation to produce the
% final score EPDMS .
\noindent\textbf{Close-Loop (NAVSIMv2).}
NAVSIMv2~\cite{navsimv2} evaluates planners using the Extended Predictive Driver Model Score (EPDMS), which aggregates penalties and weighted subscores including no at-fault collision (NC), drivable area compliance (DAC), driving direction compliance (DDC), traffic light compliance (TLC), ego progress (EP), time to collision (TTC), lane keeping (LK), history comfort (HC), and extended comfort (EC). It also introduces a \textit{Navhard} split with a two-stage protocol: (i) the planner is scored on real observations, and (ii) it is re-evaluated under perturbed future observations synthesized via 3D Gaussian Splatting around the stage-1 endpoint. The final EPDMS is obtained by Gaussian-weighted aggregation of the two-stage scores.

\noindent\textbf{Open-Loop.} For the nuScenes dataset \cite{nuscenes}, Adv-nuSc \cite{xu2025challenger},  Turning-nuScenes \cite{momad}, and nuScenes-C \cite{zhujun_benchmarking}, we employ conventional $L_{2}$ error and CR (collision rate) as metrics, aligning with the computation method of SparseDrive~\cite{sun2024sparsedrive}.

\subsection{Implementation Details}
To demonstrate the generalization capability of \textbf{GraphWorld}, we evaluate it against several strong E2E-AD baselines, including \textbf{SparseDrive}~\cite{sun2024sparsedrive} and \textbf{TransFuser}~\cite{TransFuser}.

On the \textbf{Bench2Drive}~\cite{jia2024bench2drive}, \textbf{nuScenes}~\cite{nuscenes}, and \textbf{Adv-nuScenes}~\cite{xu2025challenger} datasets, we compare GraphWorld with \textbf{SparseDrive}. For nuScenes and Adv-nuScenes, we adopt a ResNet-50~\cite{resnet} backbone with an input resolution of $256 \times 704$. The detection module operates within a circular range of 55 meters, while the online mapping module covers a $60 \times 30$ meter area (longitudinal $\times$ lateral). The motion prediction module generates 6 trajectory modes.

For \textbf{Bench2Drive}, we use a ResNet-50 backbone with 6 decoder layers and an input resolution of $640 \times 352$. We define fixed numbers of hybrid task queries, including 900 agent queries, 100 map queries, and 480 planning queries, following the standard benchmark configuration.

On the \textbf{NAVSIM}~\cite{navsimv1} dataset, we adopt \textbf{TransFuser} as the baseline and follow the official Navtrain split for training. For fair comparison, we use the same perception modules and a ResNet-34 backbone as in TransFuser.
All experiments are conducted on 8 NVIDIA RTX 4090 GPUs. We use a batch size of 48 for nuScenes and Bench2Drive, and 64 for NAVSIM. 

The models are trained for 20 epochs on nuScenes ($\sim$10.2h), 2 epochs on Bench2Drive ($\sim$49.2h), and 100 epochs on NAVSIM ($\sim$3.1h). 
We adopt the AdamW optimizer with a learning rate of $3\times10^{-4}$ and weight decay of $10^{-3}$, together with a cosine annealing schedule and linear warmup.
The training objective consists of standard multi-task losses for detection, mapping, motion prediction, and planning. 
We apply common data augmentations, including resizing, cropping, flipping, and photometric distortion, along with BEV-consistent transformations. 
Additional training techniques include gradient clipping (max norm 25), FP16 training, and a temporal queue length of 4.

% On the nuScenes dataset, we use a total batch size of 48 and train for 20 epochs, resulting in an overall training time of 10.21 hours. On Bench2Drive, the total batch size is also 48, with 2 training epochs and an overall training time of 49.21 hours. On NAVSIM, we use a total batch size of 64 and train for 100 epochs, with an overall training time of 3.09 hours.

\subsection{Main Results}
% We evaluate GraphWorld across three diverse autonomous driving benchmarks, nuScenes, Bench2Drive, and NAVSIM. For space considerations, we provide extensive qualitative visualizations of the generated trajectories in Appendix \ref{sec:Qualitative}

\noindent \textbf{Long-Horizon Planning on nuScenes (Open-Loop).} Table~\ref{tab_nuscenes_planning_6s} reports open-loop long-horizon planning results on the nuScenes validation set. While existing E2E-AD methods perform competitively at short horizons, their errors and collision rates increase significantly as the planning horizon extends. In contrast, GraphWorld consistently outperforms all baselines across all future timestamps, achieving the lowest average L2 error of 1.34m. More importantly, it reduces the average collision rate to 0.70\%, corresponding to over 22\% relative improvement over the strongest prior method. The advantage becomes more pronounced at longer horizons (5–6s), demonstrating GraphWorld’s superior long-term interaction modeling and safety-aware planning capability.

\noindent\textbf{Bench2Drive (Closed-Loop).}
Table~\ref{tab_b2d} reports results on Bench2Drive. 
GraphWorld achieves the best closed-loop performance, significantly improving Driving Score (DS) and Success Rate (SR) over prior methods. 
Compared with SparseDrive, it increases DS from 44.54\% to 51.55\% and SR from 16.71\% to 25.47\%, while also improving efficiency and comfort, indicating more stable driving behavior. 
It further shows strong advantages in interaction-critical scenarios such as merging, overtaking, and emergency braking, demonstrating effective multi-agent interaction modeling. 
Although open-loop improvements are moderate, the substantial closed-loop gains highlight the importance of interaction and long-horizon modeling beyond trajectory fitting. 
Overall, GraphWorld provides more balanced improvements across accuracy, safety, and decision-making quality.

\noindent\textbf{NAVSIMv1 navtest (Closed-Loop).}
Table~\ref{tab_NAVSIMv1} reports closed-loop planning results on the NAVSIM v1 navtest split. 
While recent E2E-based and VLA-based methods achieve strong performance, GraphWorld remains highly competitive across all metrics. 
In particular, it achieves 99.0 in NC and 97.1 in DAC, matching or surpassing most prior world-model-based approaches, while maintaining strong performance in safety-critical metrics such as TTC and EP. 
Compared with existing world-model-based methods, GraphWorld consistently improves interaction-related performance, achieving higher EP (83.2) and PDMS (90.1), indicating better decision quality and robustness in complex driving scenarios. 
Although some VLA-based methods obtain slightly higher scores on individual metrics, GraphWorld provides a more balanced performance across safety, interaction, and planning stability. 
Overall, these results demonstrate that GraphWorld effectively captures multi-agent dynamics and enables robust closed-loop planning, achieving competitive performance while maintaining strong generalization and consistency across metrics.

\noindent\textbf{NAVSIMv2 navhard (Closed-Loop).}
Table~\ref{tab_navsimv2_navhard} reports closed-loop results on the challenging NAVSIMv2 navhard split. 
While existing E2E-based and VLA-based methods achieve strong performance in Stage 1, their performance drops significantly in Stage 2, reflecting the difficulty of long-horizon and interaction-heavy scenarios. 
In contrast, GraphWorld remains highly competitive and achieves the best overall performance with an EPDMS score of 53.6, outperforming all prior methods by a clear margin.
More importantly, GraphWorld maintains strong performance across safety-critical and interaction-related metrics, including EP, TTC, and DAC, demonstrating its robustness in complex driving situations. 
Compared with prior world-model-based methods, it consistently improves both stability and decision quality, especially under long-horizon closed-loop evaluation. 
The advantage becomes more pronounced in Stage 2, where accumulated errors and interaction complexity are more severe, highlighting the effectiveness of our agent-centric relational world modeling and flow-based refinement. 
These results demonstrate that GraphWorld enables more robust long-horizon planning and better handles complex multi-agent interactions in challenging scenarios.

\noindent\textbf{NAVSIMv2 navtest (Closed-Loop).}
Table~\ref{tab_navsimv2_navtest} reports results on the NAVSIMv2 navtest split. 
While recent E2E-based and VLA-based methods achieve strong performance, GraphWorld remains highly competitive and achieves the best overall performance with an EPDMS of 89.5. 
Compared with prior world-model-based methods, it slightly improves over Latent-WAM (89.3) while maintaining balanced performance across all metrics.
Notably, GraphWorld achieves strong results on safety-critical and interaction-related metrics such as DAC and TTC, demonstrating robust decision-making in complex driving scenarios. 
Although some methods obtain higher scores on individual metrics, GraphWorld provides more consistent performance across safety, stability, and interaction dimensions. 
These results indicate that GraphWorld effectively captures multi-agent dynamics and enables reliable closed-loop planning.

\noindent \textbf{Short-Horizon Planning on nuScenes (Open-Loop).}
Table~\ref{tab_nuscenes_planning_3s} reports 3s open-loop planning results on nuScenes val. GraphWorld achieves competitive performance, with 0.57m average L2 error and a 0.08\% collision rate, on par with SOTA methods.

% This shows that GraphWorld preserves strong short-horizon performance while supporting long-horizon improvements.

\noindent\textbf{Motion Results on NuScenes.} In Table~\ref{tab_nuscenes_motion}, GraphWorld achieves the best overall motion prediction performance on nuScenes, reducing minADE from 0.62 to 0.55 (11.3\%) compared with SparseDrive. Although VAD yields the lowest miss rate, GraphWorld achieves the highest EPA, suggesting more interaction-aware multi-modal predictions.

\subsection{Robustness Study}

\noindent \textbf{Adv-nuSc (Open-Loop).} 
Table~\ref{tab_advnusc_planning} reports robustness under adversarial driving scenarios. Compared to prior methods, GraphWorld consistently achieves the lowest collision rates across all horizons, reducing the average collision rate to 0.738\%. These results demonstrate GraphWorld’s superior robustness and stability when facing aggressive and highly interactive driving behaviors.

\noindent \textbf{NuScenes-C.}
We evaluate robustness on \textbf{nuScenes-C}~\cite{zhujun_benchmarking}, a corrupted benchmark derived from the nuScenes validation set with \textbf{27 corruption types} at \textbf{5 severity levels}. Focusing on three representative adverse-weather corruptions (\textbf{Snow}, \textbf{Rain}, and \textbf{Fog}) in the open-loop setting (Table~\ref{tab_nuScenes_C_appendix}), \textbf{GraphWorld} achieves the best overall robustness, yielding the lowest average collision rates of \textbf{0.17/0.16/0.17} under Snow/Rain/Fog, respectively, substantially improving over the SparseDrive baseline (Avg.\ \textbf{0.30/0.31/0.36}). The advantage becomes more pronounced as the horizon increases, where GraphWorld maintains lower 3s collision rates (\textbf{0.28/0.29/0.29}) than all baselines, indicating stronger resistance to error accumulation under weather-induced perception degradation. Meanwhile, on the \textbf{Clean} split, GraphWorld remains competitive (Avg.\ \textbf{0.08}) and achieves the best short-horizon safety (1s/2s: \textbf{0.00}/\textbf{0.04}), suggesting that the robustness gains under corruptions are obtained without sacrificing performance in normal conditions.

\noindent \textbf{Turning-nuScenes.}
We further evaluate planning robustness on \textbf{Turning-nuScenes}~\cite{momad}, a challenging subset of nuScenes curated to emphasize \textbf{non-trivial turning maneuvers} and trajectory temporal coherence. Unlike the original nuScenes validation set that is dominated by go-straight commands, Turning-nuScenes selects samples whose GT ego trajectory exhibits a large directional change, implemented via a \textbf{25 m displacement threshold} between the GT positions at \textbf{0.5s} and \textbf{3.0s}. This yields a compact but difficult validation set with \textbf{680 samples across 17 scenes}, specifically designed to stress-test planners under high-curvature, interaction-heavy motions. As shown in Table~\ref{tab_turninguscenes_planning_appendix}, \textbf{GraphWorld} achieves the lowest collision rate across all horizons and the best overall average (\textbf{0.28}), outperforming SparseDrive (Avg.\ 0.40), DiffusionDrive$^{*}$ (Avg.\ 0.34), and MomAD (Avg.\ 0.32). Notably, the gain becomes increasingly significant at longer horizons, where GraphWorld reduces the 3s collision rate to \textbf{0.72} compared with 0.98/0.85/0.79 of the baselines, indicating stronger resistance to error accumulation in turning trajectories. These results suggest that explicitly modeling ego-centric interactions and maintaining temporally consistent world states benefits planning stability in turning-dominant scenarios, where one-shot prediction is more prone to drift and unstable control.

\begin{figure*}[t]
\centering
\includegraphics[width=1.0\linewidth]{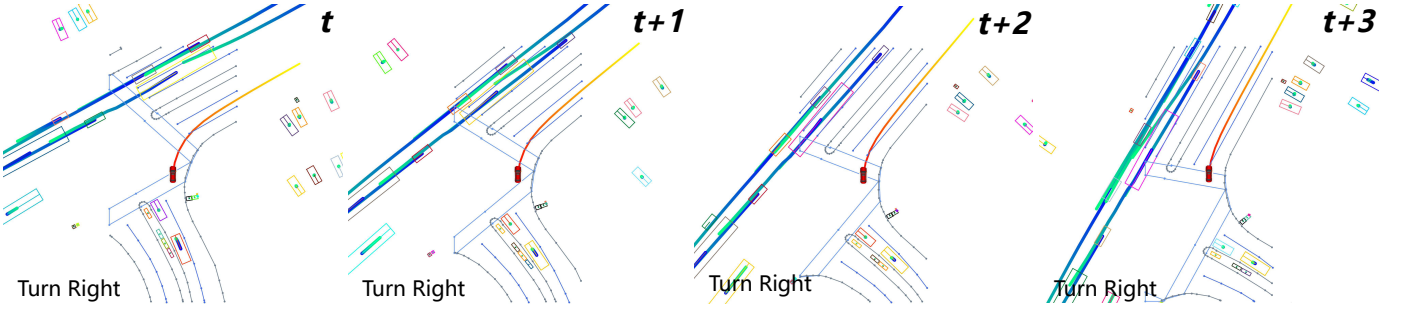}
\caption{\textbf{Visualization of GraphWorld’s 6-second long-horizon planning on the nuScenes dataset. }}
\label{fig:vis_nus_6s}
\end{figure*}

\begin{figure*}[t]
\centering
\includegraphics[width=1.0\linewidth]{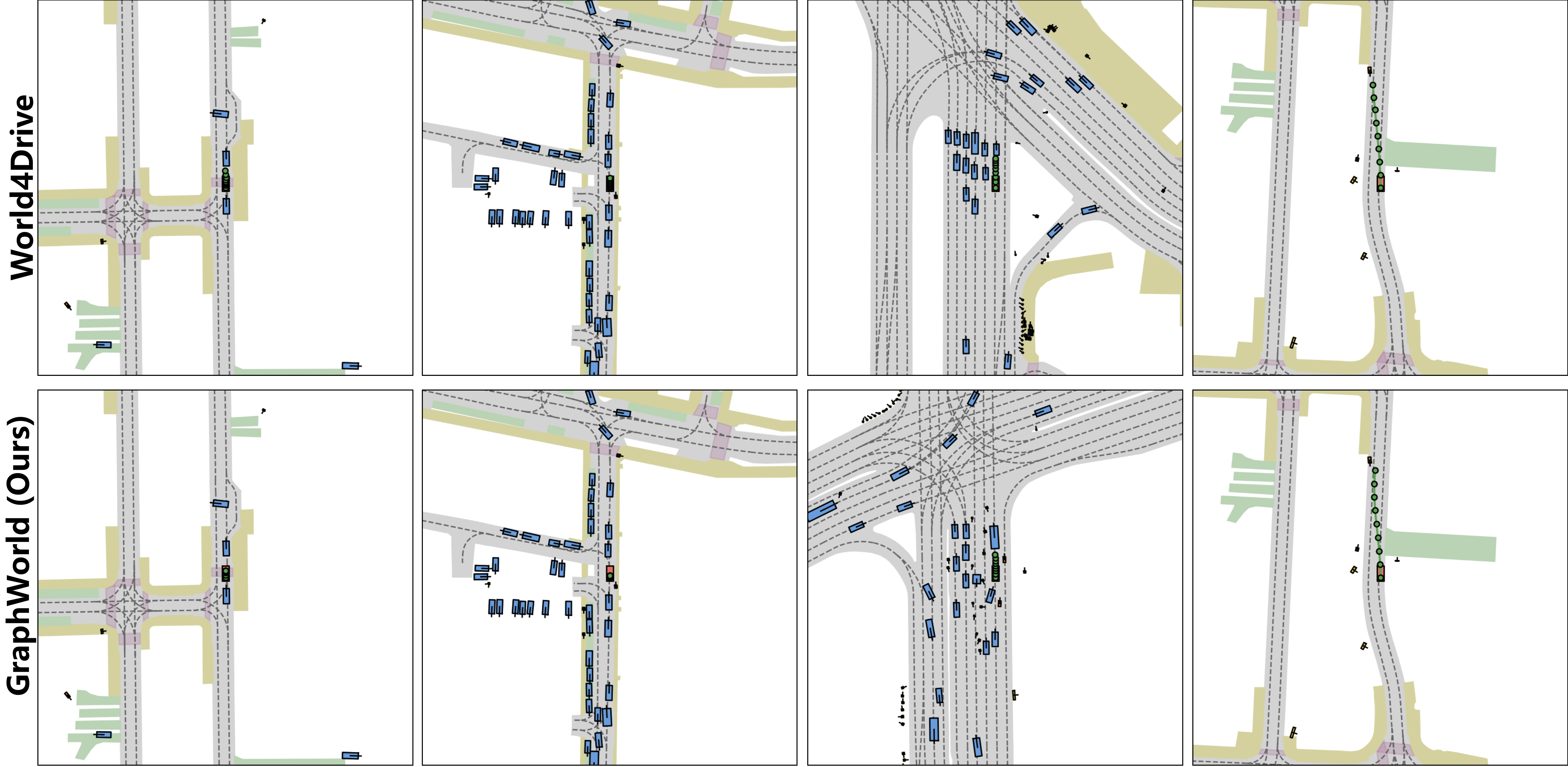}
\caption{\textbf{The visualization of GraphWorld on NAVSIMv1 dataset. }} % 和world4drive的可视化对比，我们的方法更安全，场景(VIII)和场景（I）我们的方法有效保持了和前车的间距，从而避免了碰撞，在保持行驶区域上更优，例如场景(III)和场景（IV）在其他场景上，我们的方法同样保持优势，行驶更为平滑，如场景（II）的变道场景
\label{fig:vis_navsim}
\end{figure*}

\subsection{Ablation Study}
\noindent\textbf{Roles of Different Modules in GraphWorld.}
Table~\ref{tab_ablation_modules_navsim_nuscenes_b2d} shows that both ECIG and WSCP contribute consistently to performance improvements across all datasets. 
ECIG alone already improves both open-loop and closed-loop metrics, reducing long-horizon L2 errors and collision rates on nuScenes while increasing NAVSIMv1 PDMS and Bench2Drive scores. 
Further incorporating WSCP leads to the largest gains, achieving the best overall performance across all benchmarks. 
In particular, the full model attains the highest NAVSIMv1 PDMS (90.1) and significantly improves Bench2Drive DS and SR, demonstrating enhanced decision-making and stability. 
These results highlight the complementary roles of ECIG in interaction modeling and WSCP in world-state refinement.

% As shown in Tables~\ref{tab_Ablation_navsim_different_modules} and~\ref{tab_Ablation_nuscenes_different_modules}, the ablation results on NAVSIMv1 and nuScenes consistently demonstrate the complementary roles of ECIG and WSCP. Introducing ECIG alone yields steady improvements over the baseline, reducing L2 errors (e.g., from 2.95,m to 2.76,m at 6,s) and collision rates, indicating more accurate and interaction-aware motion modeling. Incorporating WSCP on top of ECIG leads to further gains, achieving the lowest long-horizon errors (2.29,m at 6,s) and collision rates, while significantly improving planning-related metrics (e.g., NC and PDMS). These results confirm that ECIG provides structured interaction cues, whereas WSCP effectively exploits the inferred world state for safety-aware long-horizon planning.

% Table~\ref{tab_ablation_modules_navsim_nuscenes} shows that both ECIG and WSCP contribute consistently across closed- and open-loop settings. ECIG alone yields moderate gains, while combining ECIG with WSCP leads to the largest improvements, notably reducing nuScenes L2@6s from 2.95 to 2.29 and collision rate from 2.33\% to 1.95\%, demonstrating the effectiveness of world-state-conditioned long-horizon planning.

\noindent\textbf{Ablation on Different Graph Structures in ECIG.}
As shown in Table~\ref{tab_ablation_graph_structure_navsimv1_nuscenes}, the ego-centric star graph (ego→nbrs) consistently outperforms the fully-connected design across both NAVSIMv1 and nuScenes. 
It significantly improves NC from 96.1 to 99.0 and PDMS from 85.0 to 90.1, while also reducing long-horizon L2 errors and collision rates. 
These results suggest that focusing on ego-centered interactions effectively captures critical dependencies, while avoiding noisy neighbor–neighbor connections and improving overall stability.

\noindent \textbf{Ablation on Diffusion vs. Flow-Matching.}
Table~\ref{tab_all_datasets_diffusion_vs_flow} compares Diffusion and Flow-Matching across multiple datasets. 
Flow-Matching consistently outperforms Diffusion in both open-loop and closed-loop settings. 
On nuScenes, it reduces long-horizon L2 errors from $1.79/2.07/2.46$ to $\mathbf{1.64/1.88/2.29}$ and lowers collision rates from $0.75/1.45/2.23$ to $\mathbf{0.65/1.29/1.95}$, with improvements maintained at longer horizons, indicating better stability against error accumulation. 
On Bench2Drive, it improves the driving score from $73.22$ to $\mathbf{76.71}$ and success rate from $40.74\%$ to $\mathbf{44.22\%}$, while also enhancing efficiency and comfort, suggesting safer and smoother behaviors in interactive scenarios. 
On NAVSIMv1, Flow-Matching further improves all key metrics, including NC ($98.0 \rightarrow \mathbf{99.0}$), DAC ($96.6 \rightarrow \mathbf{97.1}$), and PDMS ($88.1 \rightarrow \mathbf{90.1}$). 
These consistent gains across diverse benchmarks demonstrate that Flow-Matching provides stronger robustness and generalization for trajectory generation.

\noindent\textbf{Ablation on Neighbor Selection in ECIG.}
Table \ref{tab_Ablation_nuscenes_neighbor_selection_strategies} shows that an ego-centric, distance-based neighbor selection with a moderate radius (5m) achieves the best trade-off, reducing L2 error from 2.35m to 2.29m at 6s and collision rate from 2.02\% to 1.95\%. Both overly sparse and overly dense neighbor sets degrade long-horizon accuracy and safety. 

\noindent\textbf{Ablation on Training Stages for World-State Learning.}
Table~\ref{tab_nuscenes_planning_6s_different_stages} shows that introducing the second training stage consistently reduces long-horizon L2 errors and collision rates on nuScenes, indicating that explicit temporal supervision effectively improves the stability and quality of the learned world state.

\noindent \textbf{Ablation Study of Different Times on the nuScenes Set.}
Table~\ref{tab_nuscenes_planning_6s_different_times_appendix} investigates the impact of using different temporal stages (\emph{$t{+}1$, $t{+}2$, $t{+}3$}) for long-horizon planning on nuScenes. We observe a clear trend that earlier stages consistently achieve better planning quality and safety. Specifically, $t{+}1$ yields the lowest L2 errors at all horizons ($\mathbf{1.64/1.88/2.29}$ at $\{4\text{s},5\text{s},6\text{s}\}$) and the lowest collision rates ($\mathbf{0.65/1.29/1.95}$), while performance gradually degrades as the stage shifts to $t{+}2$ and $t{+}3$. This degradation is more evident for longer horizons (e.g., 6s L2: $2.29 \rightarrow 2.44 \rightarrow 2.57$; 6s collision: $1.95 \rightarrow 2.09 \rightarrow 2.26$), indicating that later-stage rollouts suffer from increased uncertainty and error accumulation. Overall, these results suggest that leveraging earlier temporal information provides more accurate and safer trajectory generation for long-horizon planning.

\noindent \textbf{Ablation on Sampling Steps.}
Table~\ref{tab_sampling_steps} shows that increasing sampling steps yields only marginal gains. 
PDMS improves slightly from 88.3 to 89.3, while performance largely saturates after 2 steps. 
In contrast, inference speed drops significantly (51 FPS to 42 FPS). 
Therefore, we adopt 2 steps as a better trade-off between accuracy and efficiency.

\noindent \textbf{Ablation on Solver Choice.}
Table~\ref{tab_solver_choice} shows that Heun provides only marginal improvements over Euler (e.g., PDMS: $90.1 \rightarrow 89.2$), while reducing inference speed (51 FPS to 46 FPS). 
This indicates that the learned dynamics are sufficiently smooth, making Euler a more efficient and practical choice.

\subsection{Visualizations}
\label{sec:Qualitative}
% We provide qualitative visualizations of GraphWorld on  \textbf{nuScenes},  and \textbf{NAVSIMv1} to intuitively illustrate its planning behavior under different scene distributions and evaluation protocols.

\noindent \textbf{Visualization of GraphWorld’s 6-second long-horizon planning on the nuScenes dataset.} As shown in Fig.~\ref{fig:vis_nus_6s}, on \textbf{nuScenes} we visualize 6-second long-horizon planning. 
GraphWorld produces longer-range and temporally coherent trajectories, maintaining stable headings and smooth curvature variations in complex segments such as turning maneuvers. 
This capability stems from the ECIG, which explicitly captures key relations between the ego vehicle and surrounding traffic participants, together with a consistent latent world state that encodes multi-agent dynamics, thereby providing more reliable conditioning information for 6-second planning.

\noindent \textbf{Visualization of GraphWorld on  NAVSIMv1  dataset.}   Using the official implementation of the representative latent world model approach World4Drive, we visualize its behavior on NAVSIMv1 and conduct a qualitative comparison, as shown in Fig.~\ref{fig:vis_navsim}. 
The results show that GraphWorld maintains a more appropriate safety distance from leading vehicles, effectively reducing potential collision risks. 
Meanwhile, its generated trajectories are better constrained within drivable areas. 
Overall, GraphWorld demonstrates stronger \textbf{long-horizon planning} capability, enabling more stable and safety-aware decision-making over extended time horizons.

\section{Conclusion}
We %proposed
propose \textbf{GraphWorld}, an end-to-end autonomous driving framework that enhances long-horizon planning via ego-centric latent world modeling. By combining an ECIG with WSCP module and a two-stage temporal supervision strategy, GraphWorld enables efficient, interaction-aware, and safety-oriented trajectory generation. Extensive experiments on both open-loop and closed-loop benchmarks demonstrate consistent improvements in long-horizon accuracy and collision reduction, validating the effectiveness of ego-centric relational world modeling for robust autonomous driving.

\noindent \textbf{Limitations and Future Work.}
While GraphWorld improves long-horizon planning, it relies on fixed neighbor selection and single-step world-state prediction. Future work will explore adaptive graph evolution and multi-step world modeling for more complex scenarios.

\section*{ACKNOWLEDGMENTS}
% This work was supported in part by the National Key R\&D Program of China (2018AAA0100302), supported by the STI 2030-Major Projects under Grant 2021ZD0201404.
This work was supported by the National Natural Science Foundation of China (NSFC) under Grants No. 62536001 (Key Program) and No. 62576026.

\ifCLASSOPTIONcaptionsoff
  \newpage
\fi

\bibliographystyle{IEEEtran}
\bibliography{egbib}
% \makeatletter
% \renewenvironment{IEEEbiography}[2][]{\relax}{}
% \makeatother
\begin{IEEEbiography}[{\includegraphics[width=1in,height=1.25in,clip,keepaspectratio]{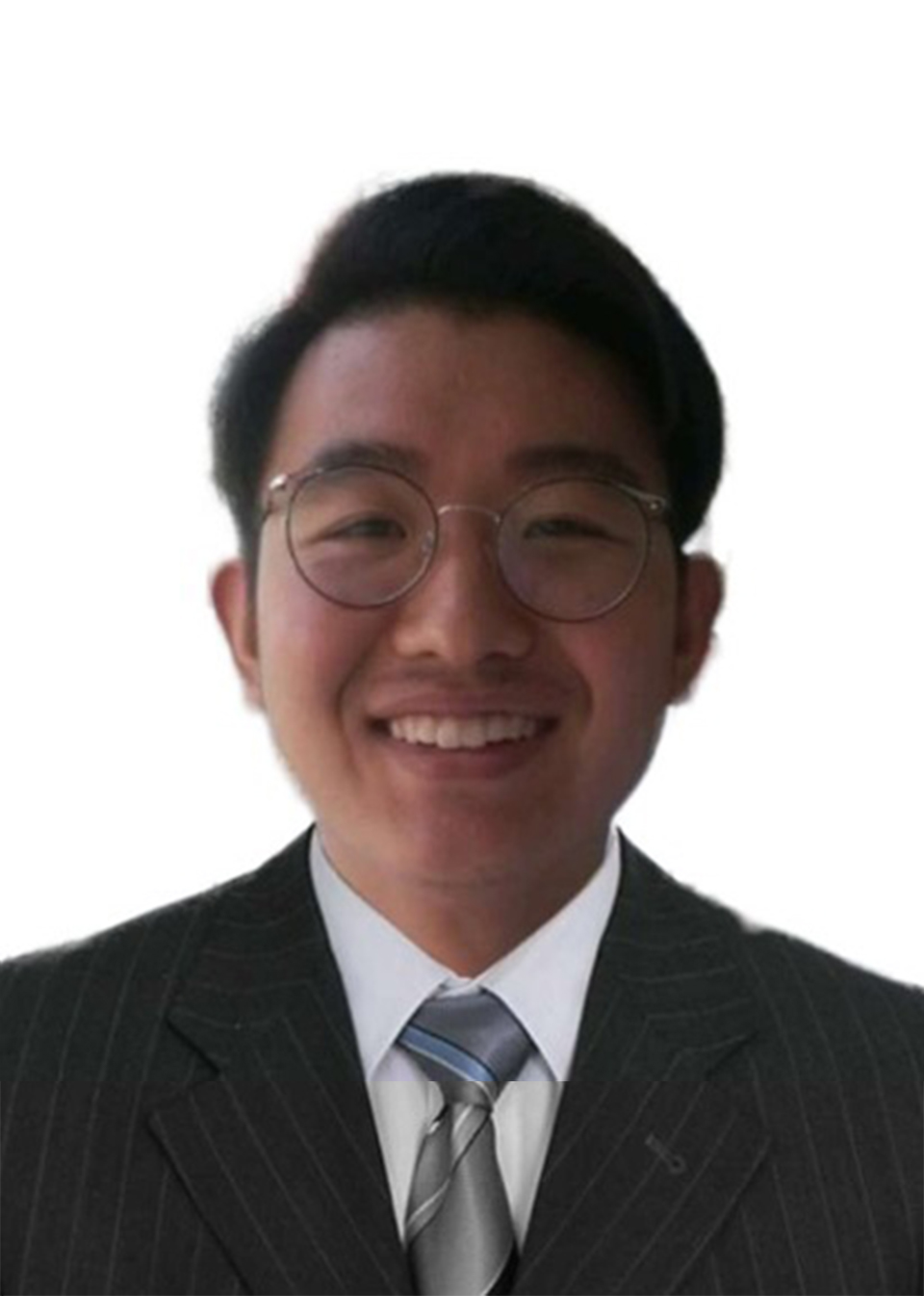}}]{Ziying Song}
 received the B.S. degree from Hebei Normal University of Science and Technology, China, in 2019, the M.S. degree from Hebei University of Science and Technology, China, in 2022, and the Ph.D. degree in computer science and technology from Beijing Jiaotong University, China, in March 2026. He is currently an Assistant Professor with the School of Artificial Intelligence, Yanshan University, China. His research interests include  autonomous driving, end-to-end autonomous driving, world models, embodied intelligence, and VLA models.
\end{IEEEbiography} \vspace{-2em}

% caiyan jia
\begin{IEEEbiography}[{\includegraphics[width=1in,height=1.25in,clip,keepaspectratio]{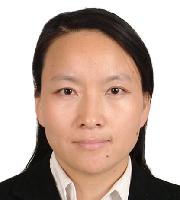}}]{Caiyan Jia}
received the B.S. degree in mathematics from Ningxia University in 1998, the M.S. degree in computational mathematics from Xiangtan University in 2001, and the Ph.D. degree in engineering from the Institute of Computing Technology, Chinese Academy of Sciences, in 2004. She is currently a Professor with the School of Computer Science and Technology, Beijing Jiaotong University, Beijing, China.
\end{IEEEbiography} \vspace{-2em}

\begin{IEEEbiography}
[{\includegraphics[width=1in,height=1.25in,clip,keepaspectratio]{{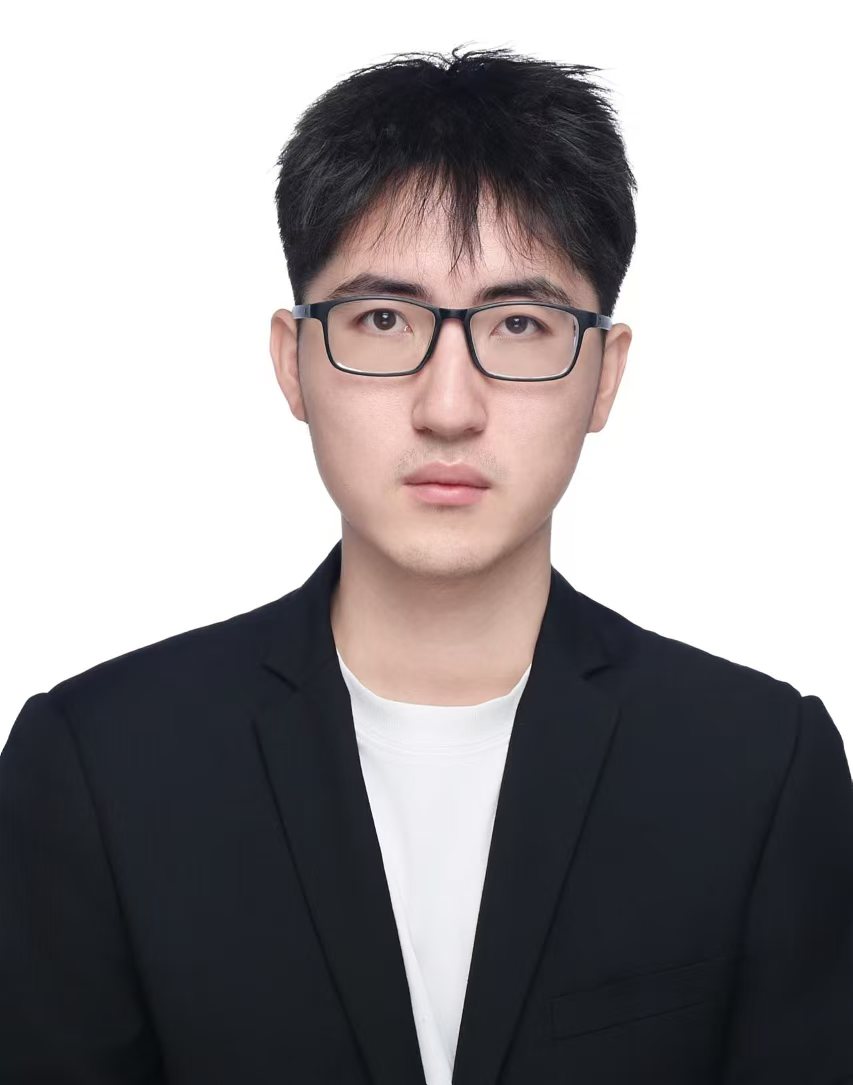}}}] {Lin Liu} was born in Jinzhou, Liaoning Province, in 2001. He received his bachelor's degree from China University of Geosciences, Beijing in 2023, and is currently pursuing his master's degree in Computer Science and Technology at Beijing Jiaotong University. Starting in September 2026, he will join Dalian University of Technology as a Ph.D. student, focusing his research on Embodied AI and Autonomous Driving.
\end{IEEEbiography} \vspace{-2em}

\begin{IEEEbiography}[{\includegraphics[width=1in,height=1.25in,clip,keepaspectratio]{{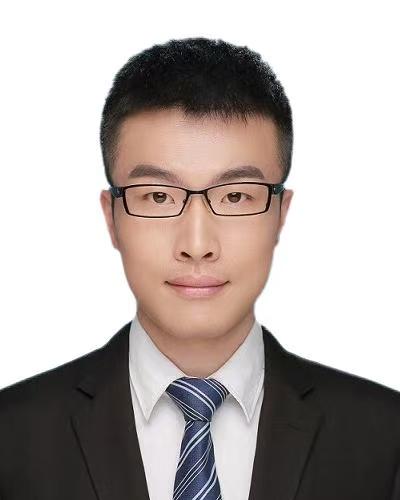}}}]
{Lei Yang (Member, IEEE)} received the M.S. degree from the Robotics Institute, Beihang University, China, in 2018, and the Ph.D. degree from the School of Vehicle and Mobility, Tsinghua University, China, in 2024. From 2018 to 2020, he joined the Autonomous Driving R\&D Department of JD.COM as an algorithm researcher. Currently, he is a research fellow with the School of Mechanical and Aerospace Engineering, Nanyang Technological University, Singapore. His current research interests include autonomous driving, 3D scene understanding and world model.
\end{IEEEbiography} \vspace{-2em}

\begin{IEEEbiography}[{\includegraphics[width=1in,height=1.25in,clip,keepaspectratio]{{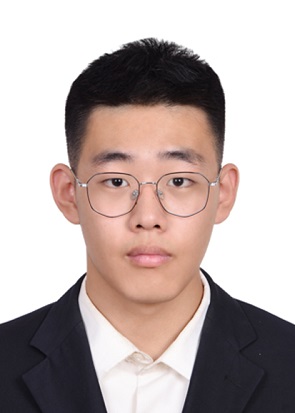}}}]
{Shengkai Zhang}  studied for the B.S. degree in Computer Science and Technology at Hebei University, China, from 2022 to 2026. Since 2026, he has been pursuing the M.S. degree in Computer Science and Technology at Beijing Jiaotong University, China. His research interests include computer vision.
\end{IEEEbiography} \vspace{-2em}

\begin{IEEEbiography}[{\includegraphics[width=1in,height=1.25in,clip,keepaspectratio]{{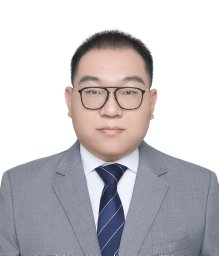}}}]
{Feiyang Jia} was born in Yinchuan, Ningxia Province, China, in 1998. He received his B.S. degree from Beijing Jiaotong University (China) in 2020. He received a master's degree from Beijing Technology and Business University (China) in 2023. He is now a Ph.D. student majoring in Computer Science and Technology at Beijing Jiaotong University (China), with research focus on Computer Vision.
\end{IEEEbiography} \vspace{-2em}

\begin{IEEEbiography}[{\includegraphics[width=1in,height=1.25in,clip,keepaspectratio]{{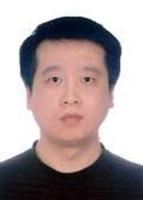}}}]
{Fengda Zhao} was born in March 1976, received his Ph.D. in Computer Software and Theory from Yanshan University. He is currently the Dean of the School of Artificial Intelligence at Yanshan University, where he also serves as a Professor and Ph.D. Supervisor. He holds additional positions as a Hebei Provincial Teaching Master, Vice Chairman of the Hebei Machine Learning Society, and a Senior Member of the China Computer Federation. His research interests include artificial intelligence applications, natural human-computer interaction, and intelligent robotics.
\end{IEEEbiography} \vspace{-2em}

\begin{IEEEbiography}[{\includegraphics[width=1in,height=1.25in,clip,keepaspectratio]{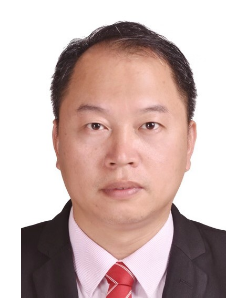}}]{Peiliang Wu} received the B.Sc. and Ph.D. degrees from Yanshan University, Qinhuangdao, China, in 2004 and 2010, respectively.  He is currently a Professor and Doctor Advisor with Yanshan University. He is a Member of the Academic Committee of Yanshan University, Member of the Standing Committee of the Youth Work Committee of the Chinese Artificial Intelligence Society, and the Vice Chairman of ACM Qinhuangdao. His research interests include robot learning and multi-agent systems.
\end{IEEEbiography} \vspace{-2em}

\begin{IEEEbiography}[{\includegraphics[width=1in,height=1.25in,clip,keepaspectratio]{{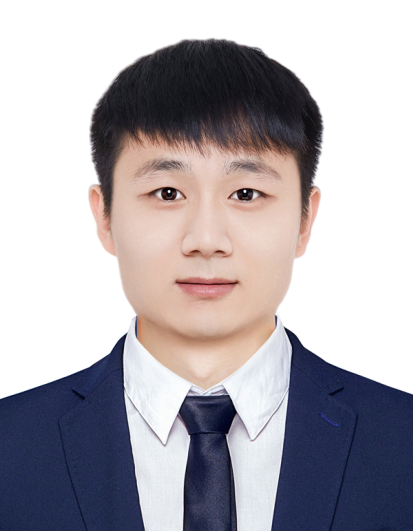}}}]{Shaoqing Xu} received his M.S. degree in transportation engineering from the School of Transportation Science and Engineering in Beihang University. He is currently working toward the Ph.D. degree in electromechanical engineering with the State Key Laboratory of Internet of Things for Smart City, University of Macau, Macao SAR, China. His research interests include 3D Space Intelligence, End2End, WorldModel, VLA, and its applications in Autonomous Driving and Robotics.
\end{IEEEbiography} \vspace{-2em}

\begin{IEEEbiography}[{\includegraphics[width=1in,height=1.25in,clip,keepaspectratio]{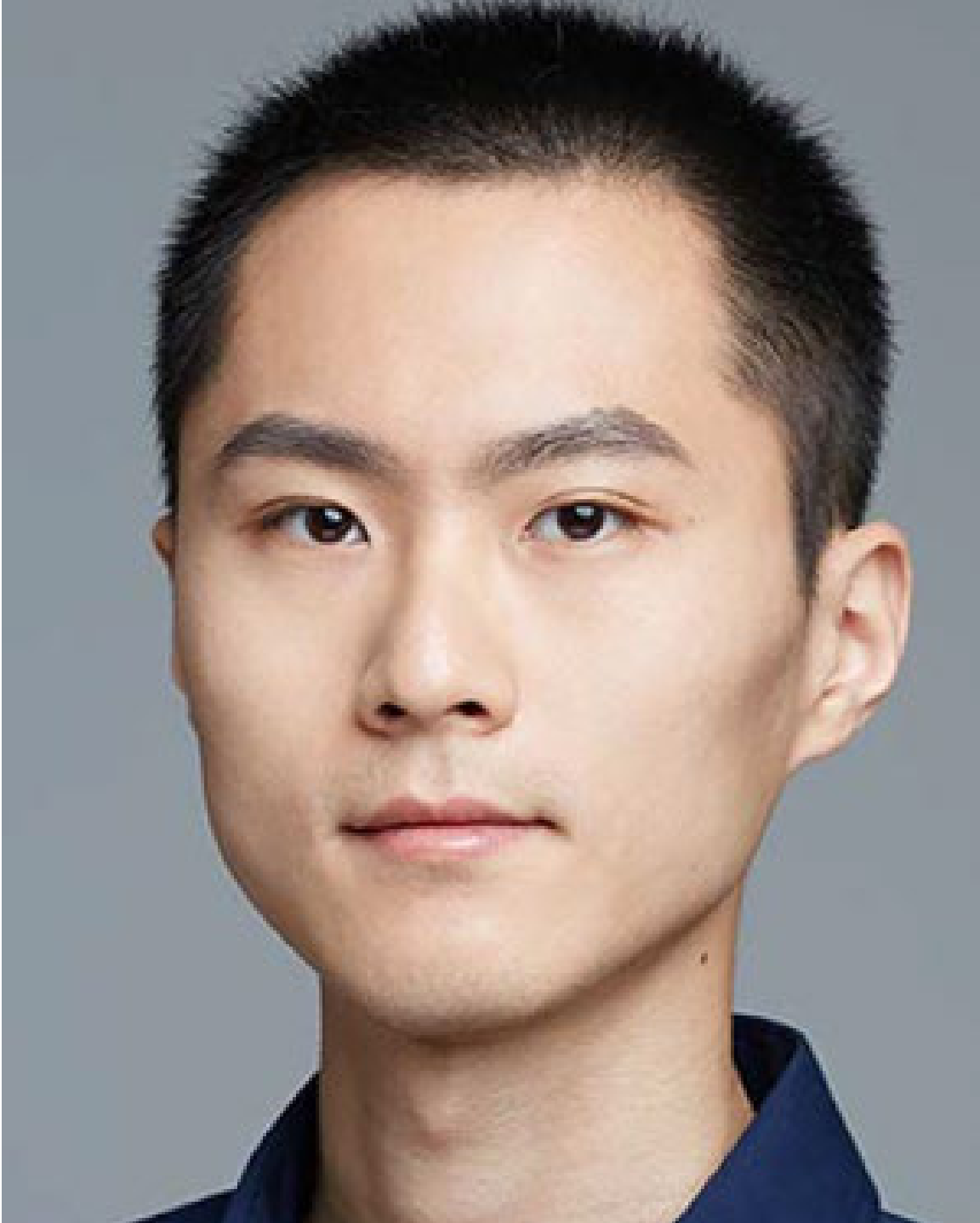}}]{Chen Lv}(Senior Member, IEEE) received the Ph.D. degree from Tsinghua University, China, in 2016. He is currently a Nanyang Associate Professor at Nanyang Technological University, Singapore, and Cluster Director of Future Mobility Solutions. His research interests include intelligent vehicles, automated driving, human–machine systems, and cyber-physical systems. He has authored two books, published over 100 papers, and holds 12 granted patents. He serves as an Associate Editor for IEEE T-ITS, IEEE T-VT, and IEEE T-IV.

\end{IEEEbiography} \vspace{-2em}

\begin{IEEEbiography}[{\includegraphics[width=1in,height=1.25in,clip,keepaspectratio]{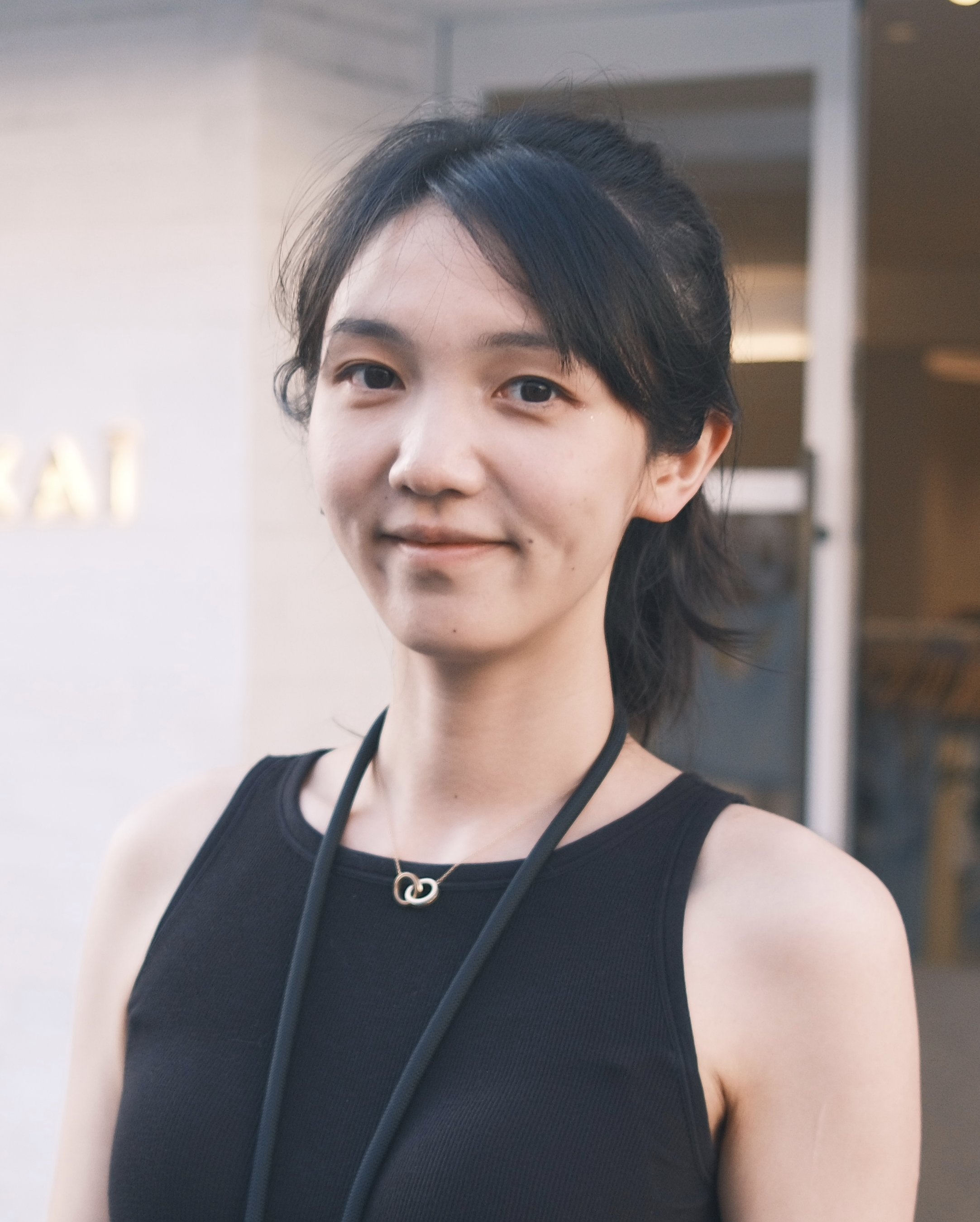}}]{Yadan Luo}(Member, IEEE) received the BS degree in computer science from the University of Electronic Science and Technology of China, and the PhD degree from the University of Queensland. Her research interests include machine learning, computer vision, and multimedia data analysis. She is now a senior lecturer and ARC DECRA at the University of Queensland.
\end{IEEEbiography} \vspace{-2em}

\end{document}

%% file: shortcuts.tex
\makeatletter
\DeclareRobustCommand\onedot{\futurelet\@let@token\@onedot}
\def\@onedot{\ifx\@let@token.\else.\null\fi\xspace}

\makeatother

\definecolor{darkgreen}{rgb}{0,0.7,0}
\definecolor{darkblue}{RGB}{31,119,180}
\definecolor{darkred}{RGB}{214,39,40}
\definecolor{mediumgray}{rgb}{0.5,0.5,0.5}
\definecolor{mediumteal}{rgb}{0,0.5,0.5}

\definecolor{ellisred}{rgb}{0.87,0.44,0.38} %
\definecolor{ellisgreen}{rgb}{0.69,0.90,0.52} %
\definecolor{elliscyan}{rgb}{0.29,0.77,0.74} %
\definecolor{ellisorange}{rgb}{0.89,0.55,0.28} %
\definecolor{ellisblue}{rgb}{0.41,0.61,0.86} %